\documentclass[11pt]{article}
\usepackage[utf8]{inputenc} 
\usepackage[T1]{fontenc}    
\usepackage{hyperref}       
\usepackage{url}            
\usepackage{booktabs}       
\usepackage{amsfonts}       
\usepackage{nicefrac}       
\usepackage{subcaption}
\usepackage{microtype}      
\usepackage[utf8]{inputenc}
\usepackage[table]{xcolor}
\usepackage{amsmath,amsbsy,amsfonts,amssymb,amsthm,dsfont,fullpage,units}

\usepackage{algorithm}
\usepackage{algorithmic}
\usepackage{adjustbox,lipsum}
\usepackage{graphicx}   
\usepackage{tabularx}  
\usepackage{authblk}
\usepackage{natbib}

\newcolumntype{C}[1]{>{\centering\let\newline\\\arraybackslash\hspace{0pt}}m{#1}}

\newcolumntype{L}{>{\centering\arraybackslash}m{5.5cm}}

\newtheorem{teo}{Theorem}[section]
\newtheorem{cor}{Corollary}[section]
\newtheorem{lemma}{Lemma}[section]

\newtheorem{defi}{Definition}[section]
\newtheorem{oss}{Remark}[section]
	
	\def\M{\bar{M}}
	
	\def\R{\mathbb{R}}
	\def\P{\mathbb{P}}
	\def\E{\mathbb{E}}

\title{A New Spectral Method for Latent Variable Models} 
\author[]{Matteo Ruffini}
\author[]{Marta Casanellas}
\author[]{Ricard Gavaldà}
\affil[]{Universitat Politècnica de Catalunya}

\begin{document}
 
\maketitle
{\def\thefootnote{}
\footnotetext{E-mail:
\texttt{matteo.ruffini@estudiant.upc.edu},
\texttt{marta.casanellas@upc.edu},
\texttt{gavalda@cs.upc.edu}}

\begin{abstract}
 This paper presents an algorithm for the unsupervised learning of latent variable models from unlabeled sets of data. We base our technique on spectral decomposition, providing a technique that proves to be robust both in theory and in practice. 
We also describe how to use this algorithm to learn the parameters of two well known text mining models: single topic model and Latent Dirichlet Allocation, providing in both cases an efficient technique to retrieve the parameters to feed the algorithm.  
We compare the results of our algorithm with those of existing algorithms on synthetic data, and we provide examples of applications to real world text corpora for both single topic model and LDA, obtaining meaningful results.
\end{abstract}

\section{Introduction}

Latent variable models (LVM) are a wide class of parametric models characterized by the presence of some hidden \textit{unobservable} variables influencing observable data. A lot of widely used models belong to this class: Gaussian Mixtures, Latent Dirichlet Allocation, Na{\"\i}ve Bayes and Hidden Markov Models and many others; in recent years, they have been object of an increasing interest in the learning literature due to the widespread of real world applications, from health-care data mining to text analytics. The huge availability of data, consequence of the development of new technologies, has boosted the need for efficient and fast algorithms to learn models belonging to this class.

Each LVM is designed as a set of observable variables (called \textit{features}) and a set of hidden variables that influence the first. Learning a LVM means, given the model structure and a sample, to infer the parameters that characterize the hidden variables and their relation with the observable features.

The classical approach was the Expectation Maximization method (EM) \citep{EMDempster}, which has been widely used  because of its generality and easiness of implementation; however EM is known to produce suboptimal results and might be very slow when the model dimension grows \citep{EMSlow}. To overcome these issues a variety of methods exploiting tensor analysis and spectral decomposition  have been recently proposed to learn various LVM, such as \citep{dasgupta1999learning,sanjeev2001learning,dasgupta2007probabilistic, vempala2002spectral, belkin2010toward, kalai2010efficiently, moitra2010settling, hsu2013learning} for mixture models or \citep{mossel2005learning,SpectralLatentHMM} for Hidden Markov Models.
In \citep{TensorLatent}, the authors presented an exhaustive survey showing that the spectral learning of most of the known LVM could be abstracted in two steps: first, given a prescribed LVM, they show how to operate with the low-order moments of the observable data in order to obtain a symmetric, low rank three-dimensional tensor; as a second step this tensor is decomposed to obtain the unknown parameters of the model. That paper accurately describes how to transform moments to obtain a symmetric tensor representation for various LVM, and  provides also one of the most popular methods to decompose the retrieved tensor and obtain the unknown model parameters: the \textit{Tensor Power Method} (TPM). TPM is an iterative technique that depends only mildly on randomized data; the main issue of this algorithm lays in its scalability, as its computational complexity depends on a factor $k^5$ where $k$ is the number of latent components. Also, TPM makes an intensive usage of tensor operations, that may be  difficult to manipulate and to understand for non-specialized practitioners (for example, a software engineer who has to maintain the code).  
A viable alternative, that in general has a better dependence on the number of latent factors, consists in dealing with matrix-based techniques, using the so called \textit{simultaneous diagonalization} approach.
Examples of these methods can be found in \citep{SpectralLatent}, where an algorithm based on the eigenvectors of a linear operator is used, and in \citep{SpectralLDA} with a method based on the singular vectors of a singular-value decomposition (SVD). These methods have a much better dependence on the number of latent states in term of complexity, but they both rely heavily on the usage of random matrices, compromising in this way the stability of the results. 
\\\\
Moving to the applications side, LVM are very popular for text mining: here the observable data, called \textit{features} of the model, is generally considered to be the words appearing in a document, while the hidden variable can be, for example, the topic of the document. A simple model for unsupervised topic mining is the single topic model, where each text is assumed to deal with a unique topic and the probability of a given word of belonging to a text depends on the topic itself of the text.  An alternative and more complex method is Latent Dirichlet Allocation (LDA) \citep[see][]{griffiths2004finding, blei2003latent}, where each text deals with more than one topic; words appear in the text according to the proportions of the topics present in the text. When using spectral methods, the standard procedure to learn these models consists in manipulating the observable moments of the data to obtain a set of symmetric, low rank tensors (examples are proposed in \citealt{TensorLatent}), and then retrieve the model parameters decomposing the retrieved tensors with a decomposition algorithm. 
\\\\
The contributions of this paper are the following:
\begin{itemize} 
\item We formally present a technique to retrieve a low-rank symmetric tensor representation for the single topic model and Latent Dirichlet Allocation. This method modifies the one presented in \citep{zou2013contrastive} increasing the robustness and the stability with respect to the noise. Also, we present a novel theorem that relates the sample accuracy of the proposed estimates to the sample size and to the lengths of the documents. 
\item We provide a new algorithm (named SVTD, \textit{\textbf{S}ingular \textbf{V}alue based \textbf{T}ensor \textbf{D}ecomposition}) to decompose the retrieved low-rank symmetric tensor. This method is  alternative to the ones presented in the cited literature, and is based on the singular values of a SVD, which are known to be stable under random perturbations (unlike the singular vectors, as shown in \citealt{SingVal}).  
Our algorithm tries to get the best from TPM and simultaneous diagonalization methods. On the one hand, as it is based itself on simultaneous diagonalization, it is simple to implement and to understand, and it scales as $k^3$ in terms of computational complexity. On the other hand, it is deterministic, not relying on any randomized matrix for its implementation; experimental results (see Section \ref{Test}) show that we reach at least the same stability of TPM, with a better scalability. The cost of this is that we require at least one feature to have different conditional expectations along the various latent states; however, we do not require the user to know in advance which this feature is, as explained in Remark \ref{selectr}. We found this requirement in general pretty natural in the real world applications: consider e.g. the topic modeling case, this requirement means that there exists at least one word whose probability of appearing is not exactly the same among the various topics. In Remark \ref{ossComp} we outline in more detail the differences between the presented method and the state of the art techniques.

\item We compare the performance of SVTD with those from the state of the art literature on synthetic data; we show that it performs at least as well as the existing methods, with a higher stability with respect to the matrix-based methods, and scales better than  TPM. Finally, we test SVTD on real world text corpora, both for single topic model and LDA, with satisfactory and meaningful results. 
\end{itemize}
 
\noindent
The outline of the paper is the following: Section \ref{singletopic} and \ref{LDASection} contain the description of the proposed technique to retrieve a low-rank symmetric tensor representation for the single topic model and a sample complexity bound; Section \ref{Sec:corealgo} contains the proposed decomposition algorithm; Section \ref{sec:pert} contains a perturbation analysis; Section \ref{Test}  tests the presented algorithm on both synthetic and real world data; Section \ref{sec:conclusion} concludes the paper outlining possible future developments and applications.
\\\\
In Sections \ref{singletopic} and \ref{LDASection} we describe a technique to retrieve a low-rank symmetric tensor representation for single topic model and for LDA; while the algorithm presented in this paper is general and can be used to learn many LVMs, it is useful to present these cases of application. 

\section{The Single Topic Model}\label{singletopic}
We consider a corpus of $N$ text documents and a set of $k$ topics; each document is deemed to belong to only one topic. The vocabulary appearing in the corpus is constituted of $n$ words, from which it is immediate to label all the words of the vocabulary with a number between $1$ and $n$. The generative process works as follows:
\begin{itemize}
\item First, a (hidden) topic $Y\in \{1,...,k\}$ is drawn, according to a given probability distribution; we define, for any $j \in \{1,...,k\}$ the probability of drawing the topic $j$ as follows:
$$
 \omega_j := \P(Y=j),\,\,\,and\,\,\,\Omega := (\omega_1,\ldots,\omega_k)'.
 $$
\item Once the topic has been chosen, all the words of the documents are generated according to a multinomial distribution; for each $i\in \{1,...,n\}$,  $\mu_{i,j}$ will be the probability of generating word $i$ under topic $j$:
$$
\P(\text{Drawing word}\, i|Y = j) = \mu_{i,j},\,\,\,and\,\,\,M = ( \mu_{i,j})_{i,j}\in\R^{n\times k}.
$$
Also we will denote with $\mu_i$ the set of columns of $M$:
$$
M = [\mu_1|,...,|\mu_k].
$$
It is a common practice to identify a topic with the probability distribution of the words under that topic, i.e. with the columns $\mu_1$,...,$\mu_k$ of $M$.
\end{itemize}
A practical encoding of the words in a document consists in identifying each word with an $n-$tuple  $x \in \R^n$, defined as:
$$
(x)_h :={\begin{cases}1&{\text{the word is}\, h},\\0&{\text{else}}.\end{cases}}
$$
In fact, if $x_j$ is the $j-$th word of a document of $c$ words, we can define $X$ as a vector whose coordinate $h$ represents the number of times the word $h$ has appeared in the document:
$$
X := \sum_{j=1}^c x_j
$$
It is common to call $X$ a \textit{bag of words} representation of a document. We can see that, if the topic is $j$, each coordinate of $X$ is distributed as a binomial distribution with parameter $c$ and $\mu_{i,j}$:
$$
Distr(X_i| Y=j) \approx B(c,\mu_{i,j})
$$
We now assume to have a corpus of $N$ documents; for each document $i\in \{1,...,N\}$ we assume to have the word-count vector $X^{(i)}$, and the total number of words in the document:
$$
c_i = \sum_{j=1}^n (X^{(i)})_j
$$
These are the only variables that we assume known, while all the parameters of the model, i.e. the pair $(M,\Omega)$, and the hidden topic of each document are supposed to be unknown.
\begin{oss}\label{ossNBMPrior}\rm
Recovering the model parameters is a useful step to infer the hidden topic of each document in a corpus. In fact, given a set of parameters $(M,\Omega)$, and a document $X$, if $Y$ is the hidden topic of $X$ we can calculate 
$$
\P(Y = j |X) = \frac{\P(X|Y=j)\,\omega_j}{\sum_{i=1}^{k}\P(X|Y=i)\,\omega_i}
$$
and assign $X$ to the topic that maximizes that probability.

\end{oss}

The following theorem is a variation of Propositions 3 and 4 in \citep{zou2013contrastive} and relates the observable moments of the known variables with the unknowns  $(M,\Omega)$. We will provide three estimators: $\tilde{M_1} \in \R^n $, $\tilde{M_2} \in \R^{n\times n} $
 and $\tilde{M_3} \in \R^{n\times n\times n} $ converging to the symmetric low rank tensors that will be used to retrieve the model parameters.
\begin{teo}\label{M123Teo}
Fix a value $N \in \mathbb{N}$, and let $ X^{(1)},...,X^{(N)}$ be $N$ sample documents generated according to a single topic model with parameters $(M,\Omega)$. The following relations hold:
\begin{itemize}
\item Define the vector $\tilde{M_1}\in \R^n$, such that for each $h\in \{1,...,n\}$:
$
(\tilde{M_1})_h := \frac{\sum_{i=1}^{N}(X^{(i)})_h }{\sum_{i=1}^{N} c_i};
$
then 
$$
\E[(\tilde{M_1})_h] = \sum_{j=1}^{k}\omega_j\mu_{h,j}.
$$
\item Define the matrix $\tilde{M_2}\in \R^{n\times n}$ such that, for each $h\neq l\in \{1,...,n\}$:
$$
(\tilde{M_2})_{h,l}  := \frac{\sum_{i=1}^{N}(X^{(i)})_h(X^{(i)})_l}{\sum_{i=1}^{N}(c_i-1)c_i}   
,\,\,\,\,\,\,
(\tilde{M_2})_{h,h}  := \frac{\sum_{i=1}^{N}(X^{(i)})_h((X^{(i)})_h-1)}{\sum_{i=1}^{N}(c_i-1)c_i};
$$
then, for each $h,l\in \{1,...,n\}$, 
$$
\E[(\tilde{M_2})_{h,l} ] = \sum_{j=1}^{k}\omega_j\mu_{h,j}\mu_{l,j}.
$$
\item Define the tensor $\tilde{M_3}\in \R^{n\times n\times n}$ such that, $h\neq l\neq m\in \{1,...,n\}$:
$$
(\tilde{M_3})_{h,l,m}  :=\frac{\sum_{i=1}^{N}(X^{(i)})_h(X^{(i)})_l(X^{(i)})_m}{\sum_{i=1}^{N}(c_i-2)(c_i-1)c_i} 
,\,\,\,
(\tilde{M_3})_{h,l,l}  :=\frac{\sum_{i=1}^{N}(X^{(i)})_h(X^{(i)})_l ((X^{(i)})_l -1)}{\sum_{i=1}^{N}(c_i-2)(c_i-1)c_i} 
$$
$$
(\tilde{M_3})_{l,l,l}  :=\frac{\sum_{i=1}^{N}((X^{(i)})_l ((X^{(i)})_l -1)((X^{(i)})_l -2)}{\sum_{i=1}^{N}(c_i-2)(c_i-1)c_i};
$$
then, for each $h,l,m\in \{1,...,n\}$, 
$$
\E[(\tilde{M_3})_{h,l,m} ] =  \sum_{j=1}^{k}\omega_j\mu_{h,j}\mu_{l,j}\mu_{m,j}.
$$
\end{itemize}
\end{teo}
Notice that, because $M_3$ ans $M_2$ are symmetric, the previous theorem defines all the entries of that operators.
Given a sample, we are able to calculate the three estimators $\tilde{M_1} $, $\tilde{M_2}  $  and  $\tilde{M_3}$. If we look at their expected values we can notice that they have a form that is highly similar to matrix products and tensor multiplications. In particular, we can express those expectations in a more synthetic form, defining the first, second and third order tensors retrieved from the observable data\footnote{We use the tensor notation to be in line with the cited literature; here, if $v_1$,...,$v_r$ are vectors  in $\R^m$,
$ v_1\otimes...\otimes v_r $ is the $m-$dimensional vector such that
$ (v_1\otimes...\otimes v_r)_{i_1,...,i_m} = (v_1)_{i_1}...(v_r)_{i_m}$.}
\begin{equation}\label{M1}
M_1 := \sum_{i=1}^{k}\omega_{i}\mu_{i} =  M \Omega
\end{equation}
\begin{equation}\label{M2}
M_2 := \sum_{i=1}^{k}\omega_{i}\mu_{i}\otimes\mu_{i} = (\sum_{j=1}^{k}\omega_j\mu_{h,j}\mu_{l,j} )_{h,l} = M diag(\Omega) M'
\end{equation}
\begin{equation}\label{M3}
M_3 := \sum_{i=1}^{k}\omega_{i}\mu_{i}\otimes\mu_{i}\otimes\mu_{i} = (\sum_{j=1}^{k}\omega_j\mu_{h,j}\mu_{l,j}\mu_{m,j})_{h,l,m}
\end{equation}
Theorem \ref{M123Teo} allows to express observable moments in the form of a symmetric tensor. By construction it is immediate to see that both $M_2$ and $M_3$ have symmetric-rank less than or equal to $k$, and the following simple limit holds, for $i = 1,2,3$:
$$
\lim_{N\to \infty}\tilde{M_i} = {M_i}
$$

We now provide a result that describes how fast this limit converges.
\begin{teo}\label{teoConv}
Let $\tilde{M_2}$ and $\tilde{M_3}$ the empirical estimates of  $M_2$ and $M_3$ obtained using Theorem~\ref{M123Teo}; define also 
$$
C_{1} = \sum_{i=1}^{N}c_i,\,\,\,\,\,\,C_{2} = \sum_{i=1}^{N}(c_i-1)c_i,\,\,\,\,\,\,C_{3} = \sum_{i=1}^{N}(c_i-2)(c_i-1)c_i
$$
$$
W_2^{(N)} = \frac{\sum_{i=1}^{N}(c_i(c_i-1))^2}{C_2^2},\,\,\,\,\,\,W_3^{(N)} = \frac{\sum_{i=1}^{N}(c_i(c_i-1)(c_i-2))^2}{C_3^2}
$$
then, for any $0\leq \delta < 1$, we have that
$$
\P (||M_2-\tilde{M_2}||_F<\epsilon)>1-\delta
$$
holds for any corpus whose document lengths $(c_1,...,c_N)$ satisfy 
$$
\sqrt{W_2^{(N)} (1 - ||M_2||_F^2)} + \sqrt{\log(\frac{1}{\delta}} )\frac{\max_j(c_j)\sqrt{C_1}}{ C_2}< \epsilon.
$$
Also, for any pair $\epsilon>0$ and $\delta>0$, we have that 
$$
P (||M_3-\tilde{M_3} ||_F<\epsilon)>1-\delta\,\,\,\,
$$
holds when
$$
\sqrt{W_3^{(N)} (1 - ||M_3||_F^2)} + \sqrt{\log(\frac{1}{\delta}} )\frac{\max_j(c_j(c_j-1))\sqrt{C_1}}{ C_3}< \epsilon
$$
\end{teo}
\begin{oss}\rm
We briefly comment on the results of the theorem. We focus on $M_2$ (similar arguments holds for $M_3$), analyzing the case where all the documents have the same length~$c$ (so, for all $i$, $c_i = c$) and $c$ is somewhat large (so $c \simeq c-1$). 
Then the bound simplifies to:
$$
\sqrt{\frac{1}{N}(1 - ||M_2||_F^2)}+ \sqrt{\frac{1}{Nc}\log(\frac{1}{\delta})}.
$$
It is interesting to notice that the worst-case accuracy of the bound is $\epsilon = O(1/\sqrt{N})$. Also, 
the bound becomes smaller as $c$ is large, with a clear limitation: if we have very few
documents ($N$ small), even if they are very long (large~$c$), it is impossible
to accurately learn the model, as in particular we may not even see all the topics.
\end{oss}
 
\begin{oss}[Alternative ways of obtaining the formulation above]\label{ossM123} \rm
The most simple technique to obtain from a text corpus described as in this section a symmetric low-rank tensor expression is the one described 
in~\citep{SpectralLatent}, that, for each document $i\in \{1,...,N\}$ considers three randomly selected words,
$
x_1^{(i)}, x_2^{(i)}, x_3^{(i)},
$
and then shows that
$$ 
\frac{\sum_{i=1}^{N}(x_1^{(i)})_h}{N}  \xrightarrow[N\to \infty]{} (M_1)_{h}
,\,\,
\frac{\sum_{i=1}^{N}(x_1^{(i)})_h(x_2^{(i)})_l}{N}  \xrightarrow[N\to \infty]{} (M_2)_{h,l}
,$$
$$
\frac{\sum_{i=1}^{N}(x_1^{(i)})_h(x_2^{(i)})_l(x_3^{(i)})_m}{N}  \xrightarrow[N\to \infty]{} (M_3)_{h,l,m}.
$$
This method is clearly unstable when dealing with small corpora, as uses only a small part of the available information (just three words for each document). 
A similar method to the one proposed here is described in \citep{zou2013contrastive}. Both estimates average the estimators with the document lengths, taking into consideration all available information; however in \citep{zou2013contrastive}, the averaging is done for each document and then they are averaged together with the same weight; for example, the off-diagonal entries of $\tilde{M_2}$, in \citep{zou2013contrastive} are calculated as follows:
$$
(\tilde{M_2})_{h,l}  :=\frac{1}{N} \sum_{i=1}^{N}\frac{(X^{(i)})_h(X^{(i)})_l}{(c_i-1)c_i} 
$$
Such calculation is a simple average of many estimators, that gives to all the documents the same weight: $\frac{1}{N}$. Instead, in Theorem \ref{M123Teo}, we propose the following different formula:
$$
(\tilde{M_2})_{h,l}  := \frac{\sum_{i=1}^{N}(X^{(i)})_h(X^{(i)})_l}{\sum_i{(c_i-1)c_i}} =\sum_{i=1}^{N} \frac{(X^{(i)})_h(X^{(i)})_l}{(c_i-1)c_i}\frac{(c_i-1)c_i}{\sum_j{(c_j-1)c_j}} 
$$
We can see that here we perform a weighted average, where the weight of the sample $i$ is $\frac{(c_i-1)c_i}{\sum_j{(c_j-1)c_j}}$, giving in practice more weight to longer documents, which are supposed to be the most reliable; we will experimentally see in Section \ref{Test} that the proposed approach is less sensitive to the noise, providing improved results. If all the documents have the same length, the two estimates will produce the same number. 
\end{oss}

\section{Latent Dirichlet Allocation}\label{LDASection}

The obvious criticism of the single topic model is that each document can deal with a unique topic, an hypothesis that is commonly considered unrealistic. To overcome this issue, more complex models have been introduced, and one of these is Latent Dirichlet Allocation (LDA) \citep{griffiths2004finding,
blei2003latent}. In its simplest form, LDA assumes that each document deals with a multitude of topics, in proportions that are governed by the outcome of a Dirichlet distribution. More precisely, considering our text corpus with $N$ documents with a vocabulary of $n$ words, the generative process for each text is the following:
\begin{itemize}
\item First a vector of topic proportions is drawn from a Dirichlet distribution with parameter $\alpha \in \R_{+}^k$, $Dir(\alpha)$; we recall that Dirichlet distribution is distributed over the symplex
$$
\Delta^{k-1} = \{v\in \R^k: \forall i, v_i\in [0,1],\,\,and\,\,\sum v_i = 1 \}
$$
and has the following density function, for $h\in \Delta^{k-1}$:
$$
\P(h) = \frac{\Gamma(\alpha_0)\prod_{i=1}^{k}h_i^{\alpha_i-1}}{\prod_{i=1}^{k}\Gamma(\alpha_i)}
$$
Where $\alpha_0 = \sum{\alpha_i}$. 
From a practical point of view, this step consists in drawing a vector of parameters $h \in \Delta^{k-1} $ such that $h_i$ represents the proportion of the topic $i$ in the document.
\item Once the topic proportions (also named \textit{mixture of topics}) have been designed, each word of the document is generated according to the following procedure: first a (hidden) topic of the word, say $Y\in \{1,...,k\}$, is drawn, according to the probabilities defined by $h$ (so we will have probability $h_j$ of drawing topic $j$) and then we will generate the word itself according to a multinomial distribution; for each $i\in \{1,...,n\}$,  $\mu_{i,j}$ will be the probability of generating the word $i$ under the topic $j$:
$$
\P(\text{Drawing word}\, i|Y = j) = \mu_{i,j},\,\,\,and\,\,\,M = ( \mu_{i,j})_{i,j}\in\R^{n\times k}.
$$
Again, we will denote with $\mu_i$ the set of columns of $M$:
$$
M = [\mu_1|,...,|\mu_k]
$$
\end{itemize}
Maintaining the notation of the previous section, we will indicate $x^{(i)}_j\in \R^n$ as the coordinate vector indicating the word at position $j$ in document $i$,  $X^{(i)} = \sum x_i$ as the word-count vector of document $i$ and $c_i =  \sum_{j=1}^n (X^{(i)})_j$ as the number of words in that document. In the case of LDA, the unknown model parameters are $M$ and $\alpha$, so the pair $(M,\alpha)$.

As in the case of the single topic model, we want to manipulate the observable moments in order to obtain a set of symmetric low rank tensors expressible as a product of the unknown parameters (as in eq. \eqref{M1},\eqref{M2} and \eqref{M3}), in order to decompose those tensors and retrieve the parameters. 
The following theorem is an immediate modification of the one presented in \citep[Lemma 3.2]{SpectralLDA}, and relates the observable moments of the known variables with the unknowns  $(M,\alpha)$; providing the required representation. The only modification consists in the fact that we have used the estimates of Theorem \ref{M123Teo} instead of the standard ones of Remark \ref{ossM123}.
\begin{teo}\label{teoLDA}
Let $\tilde{M_1}, \tilde{M_2}$ and $\tilde{M_3}$ the empirical estimates defined in Theorem~\ref{M123Teo}.
Define
$$
\tilde{M_2^{\alpha}} := \tilde{M_2} -  \frac{\alpha_0}{\alpha_0+1}\tilde{M_1} \otimes \tilde{M_1}
$$
$$
\tilde{M_3^{\alpha}} := \tilde{M_3} - \frac{\alpha_0}{\alpha_0+2}(M_{1,2}) + \frac{2 \alpha_0^2}{(\alpha_0+2)(\alpha_0+1)} \tilde{M_1} \otimes \tilde{M_1}\otimes \tilde{M_1}
$$
where $M_{1,2} \in \R^{n\times n \times n} $ is a three dimensional tensor such that
$$
(M_{1,2})_{h,l,m} = (( \tilde{M_2})_{h,l}( \tilde{M_1})_{m} + ( \tilde{M_2})_{l,m}( \tilde{M_1})_{h} + ( \tilde{M_2})_{m,h}( \tilde{M_1})_{l})
$$
 
Then
$$
\E[\tilde{M_2^{\alpha}}] = \sum_{i=1}^{k}\frac{\alpha_i}{(\alpha_0+1)\alpha_0}\mu_{i}\otimes\mu_{i} =M_2^{\alpha}
$$
$$
\E[\tilde{M_3^{\alpha}}] = \sum_{i=1}^{k}\frac{2\alpha_i}{(\alpha_0+2)(\alpha_0+1)\alpha_0}\mu_{i}\otimes\mu_{i}\otimes\mu_{i} = M_3^{\alpha}
$$
\end{teo}

\noindent
This technique allows to express observable moments in the form of a symmetric tensor. Both $M_2^{\alpha}$ and $M_3^{\alpha}$ have symmetric-rank less than or equal to $k$, and so we can use any tensor decomposition algorithm to retrieve the unknown model parameters $(M,\alpha)$ from them. A major advantage of this theorem, as of the homologous theorem in \citep{SpectralLDA}, is that it only requires the knowledge of the value  $\alpha_0$, while non-spectral methods require the knowledge of the full vector $\alpha$.

\begin{oss}[Inference]%
\label{remark:inference}%
\rm
Similarly to the single topic model, one of the main usages of LDA is to infer the mixture of hidden topics of each document in a corpus. Unfortunately, an exact formula to perform this inference is not known, but a number of approximate approaches exist, like Gibbs sampling \citep{griffiths2004finding,newman2009distributed} and Expectation Propagation \citep{blei2003latent}. In our case, if we assume to know the values of model parameters $(M,\alpha)$, we can apply a modified Gibbs Sampling to infer the topic mixture for a given text; consider a text, whose words are $x_1,...,x_c$; then, in LDA, each word $x_i$ is generated by a unique topic $Y_{x_i}$. Using the equations for Gibbs Sampling from \citet{griffiths2004finding}, if $Y_{x_i}$ is the hidden topic of word $x_i$ and $Y_{-x_i}$ is the set of topic assignment for all the words in the document excluded $x_i$, it can be shown that
\begin{equation}\label{eq:gibbs}
\P(Y_{x_i} = j| Y_{-x_i}, x_i) \approx \mu_{x_i,j} \frac{n_{-i,j} + \alpha_i}{c - 1 + \alpha_0} 
\end{equation}
\noindent
where $n_{-i,j}$ is the number of words assigned to topic $j$ excluding $x_i$, $c$ is the total number of words in the document and $\mu_{x_i,j}$ is the probability of drawing the word $x_i$ under topic $j$. So, given a document, first we have to assign to each word a hidden topic, and then update this assignment word by word in a iterative way, using a monte-carlo assignment governed by equation \eqref{eq:gibbs}. Each iteration updates the number of words assigned to a given topic; after a suitable number of iterations, we can estimate the topic mixture for a given document as the vector $h \in \R^k$ such that
$$
(h)_j = \frac{n_{j} + \alpha_i}{c  + \alpha_0} 
$$
where $n_{j}$ is the number of words assigned to topic $j$.
\end{oss}
\section{The Core Algorithm}\label{Sec:corealgo}
We now present the algorithm to retrieve the parameters of a LVM, once a symmetric tensor expression like the ones in equations \eqref{M1}, \eqref{M2} and \eqref{M3} are provided. We will use here the notation of Section 2, focusing on the single topic model, as the extension to LDA and to other LVM is straightforward~\citep[see][]{TensorLatent}.
The core of our algorithm consists in retrieving the values of the unknowns in equations  \eqref{M1},\eqref{M2} and \eqref{M3} by first getting from them a three dimensional tensor $H$ in $\R^{n\times k\times k}$ and then performing $n$ $SVD$ on the slices of $H$ (belonging to $\R^{k\times k}$) obtaining the required unknowns as the singular values of that slices; for this reason we name our method SVTD, \textit{\textbf{S}ingular \textbf{V}alue based \textbf{T}ensor \textbf{D}ecomposition}. 
We want our method to work only on with standard matrix operations,
and to accomplish this need, we select a feature (in the text mining example, a word) $r$, among the~$n$ available and, instead of considering the full tensor $M_3$, we work only with its $r-th$ slice. This is made explicit in the following definition.

\begin{defi}[Notation]\label{eq:Notation}
Given any $r\in \{1,\ldots,n\}$, we define: 
$$
 M_r :=  ( \mu_{i,j})_{i,j: i\neq r}\in\R^{(n-1)\times k},
$$
 \begin{equation}\label{CR}
 M_{2,r} := M_r diag(\Omega)   M_r' = (\sum_{j=1}^{k} \mu_{l,j}\mu_{h,j}\omega_j)_{l,h\neq r}
 \end{equation} 
  \begin{equation}\label{DR}
 M_{3,r} :=  M_r diag(\Omega)  diag((\mu_{r,1},...,\mu_{r,k})) M_r'= (\sum_{j=1}^{k} \mu_{l,j}\mu_{h,j}\mu_{r,j}\omega_j)_{l,h\neq r}
 \end{equation}  
\end{defi}

\noindent
We first observe that both matrices $ M_{2,r}$ and $  M_{3,r}$ are in $\R^{(n-1)\times (n-1)}$ and have low rank $\leq k<n$. Matrix $M_r$ is $M$ after removing the $r$-th row, as well as $M_{3,r}$ is the $r$-th slice of the three dimensional tensor $M_3$ after the removal of feature $r$.

It is easy to see that, in the case of the single topic model, all the entries of  $ M_{2,r}$ and $  M_{3,r}$ are easily estimable via the empirical formulas of Theorem \ref{M123Teo}, while Theorem \ref{teoLDA} provides us the estimates for LDA. Also for more complex LVM, we can estimate those values using the techniques outlined in \citep{TensorLatent}.

We now introduce our algorithm (SVTD), whose key steps are outlined in Algorithm~\eqref{alg3}; for a given $r$, when $M_1$, $ M_{2,r}$ and $  M_{3,r}$ are provided, this technique is able to retrieve the values of the hidden parameters $(M,\Omega)$ in few simple steps.

\begin{algorithm}[h!]
\caption{Complete algorithm - SVTD}
\label{alg3}
\begin{algorithmic}[1]
\REQUIRE $M_1,M_2,M_3$ positive semidefinite with rank $k$
\STATE Decompose $M_2$ as $M_2 = EE'$, where $E\in \R^{n\times k}$ with rank $k$
\STATE Select a feature $r$ and compute $M_{3,r}$
\STATE Compute $E_r\in \R^{n-1\times k}$ removing the $r-th$ row from $E$
\STATE Find $O$ and $(\mu_{r,1},...,\mu_{r,k})$ with a SVD on $ H_r := E_r^{\star} M_{3,r} (E_r')^{\star}$ 
\FOR{$i = 1 \to n, i \neq r$}
\STATE Compute $E_i$ removing the $i-th$ row from $E$ and get $ H_i := E_i^{\star} M_{3,i} (E_i')^{\star}$ 
\STATE Obtain $(\mu_{i,1},...,\mu_{i,k})$ as the diagonal entries of $O' H_i O$
\ENDFOR
\STATE Obtain $\Omega$ solving $M_1 = M\Omega$

\RETURN $(M,\Omega)$
\end{algorithmic}
\end{algorithm}

The constructive proof of the following theorem will explain why SVTD performs a correct retrieval of the desired model parameters. 
 \begin{teo}\label{teoalgo}
If all the elements of $(\mu_{r,1},...,\mu_{r,k})$ are distinct and $ M_{2,r}$ and $  M_{3,r}$ have rank $k$, then SVTD produces exactly the values of $(M,\Omega)$.
\end{teo}
\proof
Given $M_2$ we can decompose it as 
$$
M_2 = EE'
$$
where $E $ is a rank$-k$ matrix in $\R^{n\times k}$. If now we consider $M_{2,r}$, we can easily show that 
$$
M_{2,r} = E_rE_r'
$$
where $E_r $ is a rank$-k$ matrix in $\R^{(n-1)\times k}$ obtained by removing the $r$-th column from $E$. This decomposition is unique up to an isometry of $\R^k$; this means that there exists an orthogonal matrix $O\in \R^{k\times k}$ such that
$$
M_{2,r} = M_r (diag(\Omega)) M_r'= E_rO O'E_r'
$$
and so
\begin{equation}\label{EO}
E_rO = M_r (diag(\Omega))^{\frac{1}{2}}.
\end{equation}
We now look for that isometry, exploiting the matrix $M_{3,r}$. By construction we have
\begin{equation}\label{M3r}
M_{3,r} = M_r diag(\Omega)  diag((\mu_{r,1},...,\mu_{r,k})) M_r' = E_rO diag((\mu_{r,1},...,\mu_{r,k}))O'E_r'.
\end{equation}
So it holds
\begin{equation}\label{PINV}
H_r: = E_r^{\star} M_{3,r} (E_r')^{\star} = O diag((\mu_{r,1},...,\mu_{r,k})) O'
\end{equation}
where $E^{\star}_r = (E_r'E_r)^{-1}E_r'$ is the Moore–Penrose pseudoinverse of $E_r$.
As all the elements of $(\mu_{r,1},...,\mu_{r,k})$ are distinct, the decomposition at the right side of the equation is unique up to a reordering of the columns of $O$ and possible change of sign. $O$ can be obtained just by performing a SVD on $E^{\star}_r M_{3,r} (E_r')^{\star} $, and then used to diagonalize it as follows:
\begin{equation}\label{eq13}
 O' H_r O = diag((\mu_{r,1},...,\mu_{r,k})).
\end{equation}
In this way we retrieve the vector $(\mu_{r,1},...,\mu_{r,k})$ and the matrix $O$.
As Lemma \ref{lemmaO} says, the matrix $O$ does not depend on the feature we selected for its construction; this means that if we would have chosen a different feature $r$, the matrix $O$ would have come out to be exactly the same up to a column reordering. So, we arbitrary choose the column reordering obtained by isolating the feature $r$ and we obtain the values of the vectors $(\mu_{i,1},...,\mu_{i,k})$, for any other $i\neq r$, just by calculating 
\begin{equation}\label{eq:H}
H_i: = E_i^{\star} M_{3,i} (E_i')^{\star}  
\end{equation}
and getting $(\mu_{i,1},...,\mu_{i,k})$ from
$$
 O' H_i O = diag((\mu_{i,1},...,\mu_{i,k})).
$$
with the same $O$ used for the feature $r$. Iterating on the various features $i\in \{1,...,n\}$ we obtain the matrix $M$. 
The subsequent estimations of $\Omega$ is straightforward, and can be obtained by solving the linear system $M_1 = M\Omega$.
\endproof
As can be seen from the proof, the method just consists in three logical steps: first, we retrieve the matrix $O\in \R^{k\times k}$, then we get tensor $H\in\R^{n \times k\times k}$ whose $i-th$ slice is $H_i \in \R^{k\times k}$ defined as in \eqref{eq:H} and as a last step we get the values of the rows of $M$ as the diagonal elements of $O' H_i O$.

\begin{lemma}\label{lemmaO}
Let $E\in \R^{n\times k}$ with rank $k$ satisfying $M_2 = EE'$, and  $E_r \in \R^{(n-1)\times k}$ as $E$ with the $r-th$ row removed.
Then, there exist two isometries  $O_r$ and $O$ realizing the equations  
$$
E_rO_r = M_r (diag(\Omega))^{\frac{1}{2}},\,\,\,\,\,\,\,\,\,
EO = M (diag(\Omega))^{\frac{1}{2}}
$$
and it holds $ O = O_r$.
\end{lemma}

\begin{oss}[On the generality of the algorithm]\rm
We remark that during the construction of the algorithm we have not made any hypotheses on the probability distribution of the data; instead, we have required the matrix $M$ to be full rank, with at least one feature $r$ with different conditional expectations on the various topics. Also, we do not need to know in advance what this feature is, as Remark \ref{selectr} explains. This last requirement is not present in the other matrix-based methods, as they rely on a randomized matrix to guarantee the uniqueness of the results (see Remark \ref{ossComp}); the consequence of this dependence on randomized vectors is the introduction of  additional variance in the results. It is an interesting open problem to find a deterministic method joining the scalability properties of simultaneous diagonalization methods, without requiring this separation condition.

Regarding the latent variable models, we can state that the presented algorithm only decomposes a set of symmetric tensors, obtaining as output the unknown parameters of the model that underlies the data. A consequence is that SVTD can be used  to learn efficiently many kind of latent variable models: Gaussian mixtures, Latent Dirichlet allocation, Hidden Markov model, and all the models described in \citep{TensorLatent}. In this sense, this algorithm is a new alternative to the tensor decomposition method described in that work, or to the methods presented in \citep{SpectralLatent} or \citep{SpectralLDA}. In the experiments section we will compare the ability of this method to learn the model parameters, compared with current state of the art algorithms. 
\end{oss}

\begin{oss}[On Finding $E$]\label{ossEE}\rm
There exists a straightforward way to calculate the matrix $E$: as matrix $M_{2}$ is a symmetric positive semidefinite rank$-k$ matrix, it has a SVD:
$$
M_2 = USU' = U_kS_kU_k'
$$
where $U_k$ and $S_k$ are the matrices of singular vectors and values truncated at the $k-th$ smallest eigenvalues (the smallest non-zero eigenvalue).
Then, we can easily find the matrix
$$
E = U_kS_k^{\frac{1}{2}} \in \R^{n\times k}.
$$
\end{oss}
\begin{oss}[On the selection of feature $r$]\label{selectr}\rm

The initial steps of the algorithm need the isolation of a feature $r$ to compute the matrix $O$. While theoretically we could select any feature $r$ such that all elements of $(\mu_{r,1},...,\mu_{r,k})$ are distinct, it is clear that, if matrices $M_1$, $M_2$ and $M_3$ are subject to perturbations, the results obtained by the algorithm might vary a lot, depending on the selected feature $r$. Theorem \ref{pert} will show that the accuracy of the algorithm under perturbed data will depend on how different are the elements in $(\mu_{r,1},...,\mu_{r,k})$. A consequence of this is that a good way to find feature $r$, and so a reliable matrix $O$, is to repeat the steps 2, 3 and 4 of the algorithm, isolating different features and select the one that maximizes the quantity 
$
\min_{i\neq j}(|\mu_{r,i}-\mu_{r,j}|).
$
With this method, a user, could run SVTD without any previous knowledge on the feature to extract. This operation has an additional computational cost, as it  requires to perform $n$ times a $k\times k$ SVD; however we will see in the next remark that this cost has not a great impact on the total computational complexity, as it is dominated by the cost of other, more expensive,  computations.
\end{oss}
\begin{oss}[Complexity analysis]\label{oss:compcomplexity}\rm
We start analyzing the time complexity.
Using randomized SVD techniques \citep[see][]{RandSVD}, step 1 can be carried out with a total of $O(n^2k)$ steps. Step 4 also requires $O(n^2k)$ steps for matrix multiplication, $O(nk^2)$ steps for pseudo-inverse computation, while the SVD on $  E_r^{\star} M_{3,r} (E_r')^{\star}$ requires $O(k^3)$ steps.
The operations in the loop of steps 6 and 7 just require $O(nk^2)$ steps for the calculation of the pseudo-inverse $ E_r^{\star}$ and $O(n^2k)$ for the matrix multiplication of step 7; so the loop has total complexity
$
O(n^3k + n^2k^2).
$
The overall complexity of the Algorithm \ref{alg3}  is thus
$$
O(n^3k + n^2k^2 + k^3).
$$
To this, we should add the additional computational cost of the feature-selection method outlined in Remark \ref{selectr}: that method requires to perform $n$ times a $k\times k$ SVD, costing $O(n k^3)$; however, as $n\geq k$, this cost is dominated by the $O(n^2k^2)$ components.
It is important to highlight that the implementation described in Algorithm \ref{alg3} has mainly a descriptive purpose; for a specific LVM, optimized implementations may exists. For the single-topic model, for example, the method can be implemented without never calculating explicitly the tensor $M_3$, calculating in one step the tensor $H$, with a complexity of $O(Nnk)$ and then performing the subsequent diagonalizations  in $O(n^2k^2 )$ time.
Additional tuning of the performances can be obtained exploiting the sparsity of $X$, using for matrix operations sparse matrix technique.
We also remark that the algorithm is trivially parallelizable: assuming we have $m$ machines on which to parallelize steps 5, 6, 7 of the algorithm and the feature selection task, we can reduce the total running time to
$$
O(\frac{n^3k + n^2k^2}{m} + k^3).
$$
Regarding memory, notice that we never use the full tensor $M_3$, but only its $r-th$ slice; the overall memory complexity of the algorithm is thus $O(n^2)$.  

These complexity requirements are comparable to the ones of \citet{SpectralLatent, SpectralLDA}; however, these methods are randomized, with nontrivial variance in their output, so they may require several runs of the full algorithm in order to provide accurate results.  Tensor power method from \citet{TensorLatent} has in general a worst computational complexity: it is an iterative technique, with a  number of iterations difficult to bound a priori; the authors suggest that accuracy $\epsilon$ can be reached with $O(k^{5+\delta}(\log(k) + \log \log(1/\epsilon)))$ operations, among iterations, random restarts and actual matrix operations, compared to our $O(k^3)$; to this time we need to add the time necessary to get the $k\times k\times k$ tensor from the sample, a computational time that is not trivial for many LVMs.
\end{oss}

\begin{oss}[Alternative algorithms]\label{ossComp}\rm
As said in the introduction, other algorithms exists to retrieve the unknowns $(M,\Omega) $ from $M_1,M_2 $ and $M_3$. The most popular and solid way is TPM,  described in \citep{TensorLatent}. 
In that algorithm, the idea is to find a matrix $W$ (it may be, for example, the pseudoinverse of the matrix $E$ we retrieve at line 1 of Algorithm \ref{alg3}), such that $W'M_2W = I$, and use it to whiten the tensor $M_3$, getting a $k\times k\times k$ tensor $T$, from whose robust eigenvectors it is possible to retrieve the model parameters. To get the set of the robust eigenvectors, the authors use a three-dimensional extension of the well-known matrix power method. While very robust, the implementation of this method may result complex for who is not familiar with tensors; in addition, it is an iterative method, and so it requires a tuning of the hyperconvergence parameters, that might require many trial-and-error tests. This practical considerations, together with the high computational complexity, as outlined in Remark \ref{oss:compcomplexity}, are drawbacks that matrix methods do not have.

Matrix methods, or ''simultaneous diagonalization'' methods, as those outlined in \citep{SpectralLDA} and \citep{SpectralLatent}, are technically more similar to the methods presented here, and they are two variations of the same approach, one, that in  \citep{SpectralLatent}, using eigenvectors, and the other in \citep{SpectralLDA} using singular vectors.  
Those methods take a random vector $\eta \in \R^n$, and observe that the matrix $M_3(\eta)$, defined as 
$$
(M_3(\eta))_{i,j} := \sum_{l=1}^n (M_{i,j,l}\eta_l) 
$$
can be decomposed as follows:
$$
M_3(\eta) =  M diag(\Omega)  diag((\eta\mu_{1}',...,\eta\mu_{k}')) M'.
$$
Then, they calculate the matrix $E$ exactly as in step 1 of Algorithm \ref{alg3} and get the matrix 
$$
H_{\eta} = E^{\star}M_3(\eta)(E^{\star})'
$$
from whose left singular vectors $O$ they retrieve $M$ ( up to rescaling and columns reordering) solving 
$$
M(diag(\Omega))^{\frac{1}{2}} = EO
$$
using essentially Equation \eqref{EO}. The introduction of the random vector $\eta$ has the scope of guaranteeing that the elements of $(\eta\mu_{1}',...,\eta\mu_{k}')$ are almost surely distinct, and so $O$ is unique; the cost of this is introducing additional variance to the model, compromising the stability.
So, we can see that there are essentially two main differences: the first is the fact that instead of using a randomized matrix, we fix 
a specific feature $r$, choosing the one with the maximum minimum variation between the feature components; in particular, this is the same of saying that we fix $\eta$ to be the $r-th$ coordinate vector, providing a recipe for finding $r$ in Remark \ref{selectr}. In this way, at once, we get rid of the noise, and we provide the choice that maximizes the stability. The second difference is the fact that we do not retrieve the matrix $M$ from Equation $\eqref{EO}$, but observing that, if  $\eta_i$ is the $i-th$ coordinate vector, then 
$$
M_3(\eta_i) =  M diag(\Omega)  diag((\mu_{i,1},...,\mu_{i,k})) M'.
$$
and so, for each $i = 1,...,n$, we can find the row $(\mu_{i,1},...,\mu_{i,k})$ of $M$ as the singular values of the various $H_i = E^{\star}M_3(\eta_i)(E^{\star})'$. In this sense, our method relies on the singular {\em values} of a SVD decomposition and not on the singular vectors. In our experiments, see Section \ref{Test}, we found this approach much more stable if compared with other methods, even when their dependence on the random matrix was removed. 
\end{oss}

\section{Perturbation Analysis}\label{sec:pert}

In the previous section we have outlined an algorithm that accurately learns the model parameters of a LVM, given the exact values of $M_1$, $M_2$ and $M_3$. However, when applying our algorithm to a real world problem, we never have these exact variables, but only a set of estimators  $\tilde{M_1}$, $\tilde{M_2}$ and $\tilde{M_3}$ that are expected to become arbitrarily accurate as sample size increases.
This fact has some immediate consequences: first we need to adapt SVTD
to deal with perturbed  matrices (as  $\tilde{M_1}$, $\tilde{M_2}$ and $\tilde{M_3}$ are) 
that do not necessarily have rank~$k$ or are not assured to be positive definite.
Second, we need to study how the perturbations on those estimates propagate up to the final results.
The adaptation of SVTD is outlined in Algorithm \ref{alg2}\footnote{A python implementation of this algorithm can be found in \url{https://github.com/mruffini/SpectralMethod.git}.}. 

\begin{algorithm}[h!]
\caption{SVTD when $\tilde{M_1}$, $\tilde{M_2}$ and $\tilde{M_3}$ are available, instead of $M_1$, $M_2$ and $M_3$.}
\label{alg2}
\begin{algorithmic}[1]
\REQUIRE $\tilde{M_1},\tilde{M_2}, \tilde{M_3}$, rank $k$
\STATE Perform an SVD on $\tilde{M_2}$, obtaining $\tilde{M_2} = \tilde{U}\tilde{S}\tilde{V}'$.
\STATE Obtain $\tilde{E}\in \R^{n\times k}$ as $\tilde{E} = \tilde{U_k}\tilde{S_k}^{\frac{1}{2}}$, where $\tilde{U_k}$ and $\tilde{S_k}$ are $\tilde{U}$ and $\tilde{S}$ truncated at the $k-th$ singular vector. 
\STATE Select a feature $r$ and compute $\tilde{M}_{3,r}$
\STATE Compute $\tilde{E_r}\in \R^{n-1\times k}$ removing the $r-th$ row from $\tilde{E}$
\STATE Find $\tilde{O}$ and $(\tilde{\mu}_{r,1},...,\tilde{\mu}_{r,k})$ as the left singular vectors and values of $ \tilde{H}_r:=\tilde{E_r^{\star}} \tilde{M}_{3,r} (\tilde{E_r}')^{\star}$ 
\FOR{$i = 1 \to n, i \neq r$}
\STATE Compute $\tilde{E}_i$ removing the $i-th$ row from $\tilde{E}$ and get  $\tilde{H}_i : = \tilde{E_i^{\star}} \tilde{M_{3,i}} \tilde{(E_i')^{\star}} $
\STATE Obtain $(\tilde{\mu}_{i,1},...,\tilde{\mu}_{i,k})$ as the diagonal entries of $\tilde{O}'\tilde{H}_i\tilde{O} $
\ENDFOR
\STATE Obtain $\tilde{\Omega}$ solving $\tilde{M_1} = \tilde{M}\tilde{\Omega}$

\RETURN $(\tilde{M},\tilde{\Omega})$
\end{algorithmic}
\end{algorithm}
 
The modifications with respect to Algorithm \ref{alg3} need to guarantee 
that we deal with positive definite matrices with rank $k$. We do this by defining $\tilde{E}$, in steps 1 and 2, as the product of the first $k$ left singular vectors of $ \tilde{M}_2$ and the first $k$ singular values and, in step~5, taking $\tilde{O}$ to be the left singular vectors of $\tilde{M}_{3,r}$; the rest of the algorithm is identical.
\begin{oss}\rm
In step 1 of Algorithm \ref{alg2} we obtain
$$
\tilde{M_2} = \tilde{U}\tilde{S}\tilde{V}.
$$
In general, as $\tilde{M_2}$ converges to $M_2$ which is positive semidefinite, we expect that for suitably big samples $V = U'$, i.e. that $\tilde{M_2}$ is also positive semidefinite. The size of the sample required to have $\tilde{M_2}$  positive semidefinite depends on the concentration properties of the sample. However, Algorithm \ref{alg2} produces accurate results even when positive definiteness of $\tilde{M_2}$ is not guaranteed, as explained next in Theorem~\ref{pert}.
\end{oss}

\noindent
We now study the accuracy of the algorithm. Intuitively, the more similar the perturbed $\tilde{M_1}$, $\tilde{M_2}$ and $\tilde{M_3}$ are to the exact $M_1$, $M_2$ and $M_3$, the better the outcomes of the algorithm should be. This intuition is confirmed by the following theorem.
\begin{teo}\label{pert}
%
Given the unperturbed versions of  of $M_1$, $M_2$ and $M_3$ and the feature we want to isolate, $r$, let {$\alpha_r$ } and $\alpha_{M_2}$ be 
$$
\alpha_r = \min_{i\neq j}(|\mu_{r,i}-\mu_{r,j}|)>0,\,\,\,\,\,\,\,\alpha_{M_2}=\min_{i\leq k}{(\sigma_{i}(M_2)^2-\sigma_{i+1}(M_2)^2)}>0
$$
where $\sigma_{i}(M_2)$ are the singular values of $M_2$. Assume the empirical estimates $\tilde{M_2}$ and $\tilde{M_3}$
satisfy
$$
||\tilde{M_2}-{M_2}||_F <\epsilon,\,\,\,\,\,\,\,||\tilde{M_3}-{M_3}||_F <\epsilon. 
$$
Then, there exists a {function}\footnote{For an explicit formulation of the value of $\gamma(M,\Omega)$, we refer the reader to the proof of the theorem and to Remark \ref{oss:numberbound}.}
$
\gamma(M,\Omega)
$, of the model parameters, such that, 
if $\epsilon<\gamma(M,\Omega)$, Algorithm \eqref{alg2}, fed with $\tilde{M_1}$, $\tilde{M_2}$ and $\tilde{M_3}$, provides an estimated matrix 
$$
\tilde{M} = \begin{bmatrix}
    \tilde{m_1} \\
    \vdots\\
    \tilde{m_n}
  \end{bmatrix}
$$
whose rows satisfy, for all $i\in\{1,...,n\}$,
$$
||m_i-\tilde{m_i}||_2\leq  C_1\epsilon + \sqrt{k}\frac{\epsilon}{\alpha_r} (C_2  + \frac{C_3}{\alpha_{M_2}} )+O(\sqrt{k} C_4\epsilon^2)
$$
where $m_i$ are the rows of $M$, $C_1$, $C_2$ and $C_3$ are polynomial functions of $ ||E_i||_F,||O||_F$ and $||M_{3,i}||_F$, and $C_4$ is a polynomial function of $ ||E_i||_F,||O||_F,||M_{3,i}||_F,\frac{1}{\alpha_r}$ and $\frac{1}{\alpha_{M_2}}$.
\end{teo}

Note the key role of $\alpha_r$, the minimum difference between the elements of $(\mu_{r,1},...,\mu_{r,k})$; when samples are large enough, the theorem guarantees that the algorithm works correctly, although ``large enough'' depends on~{$\alpha_r$}. When this condition is not satisfied, the learning algorithm still works, but might provide output results that are different from the theoretical generative model. We recall here Remark~\ref{selectr}, where we wondered how to select the proper feature~$r$; ideally, the one that would guarantee the highest accuracy would be the one with the highest possible~$\alpha_r$. 

Theorem \ref{pert}, together with Theorem \ref{teoConv}, (resp. Theorem \ref{teoLDA}), provides a sample complexity bound for the Single Topic Model (resp. for LDA). For any given accuracy that ones wants to obtain in the estimates of the parameters $(M,\Omega)$, using Theorems \ref{teoConv} and \ref{pert} one can understand the sample size needed to reach that accuracy with high probability. 
\section{Experiments}\label{Test}
In this section we will provide some experiments, to test both on synthetic and real data, the algorithm we presented in this paper.

\subsection{Recovering \texorpdfstring{$M_2$}{} and \texorpdfstring{$M_3$}{}}
\begin{figure}[h!]
\centering
   \includegraphics[height=5cm]{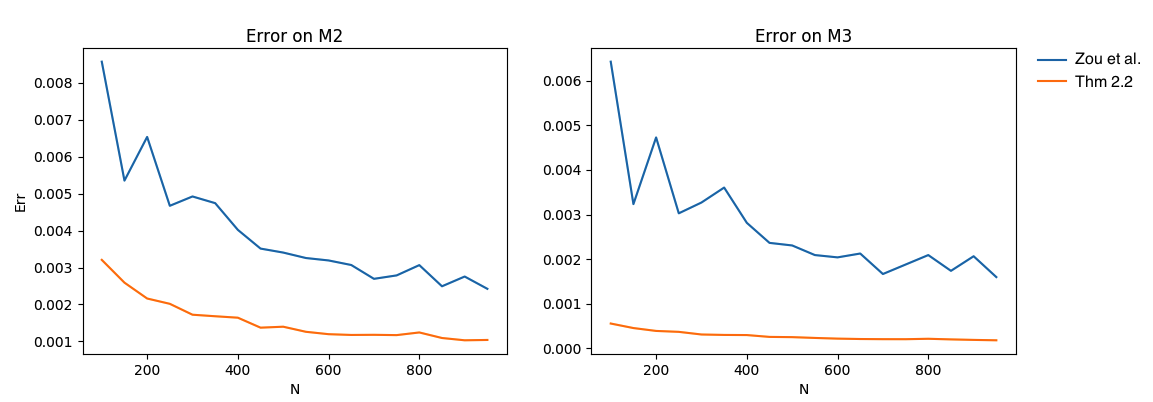}
   \label{fig:Ng1} 
\caption{The $x-$axis of the figures represents the size of the synthetic text corpora, wile the $y-$axis is $Err_2$ for the left chart and $Err_3$ for the right chart. Blue lines represent the errors obtained with the method presented in Theorem \ref{M123Teo}, while green lines refer to the method of \citet{zou2013contrastive}}\label{FigResultsM12}.
\end{figure}
In Section \ref{singletopic} we described a technique to recover the matrix $M_2$ and tensor $M_3$ from a sample, comparing it with the methods presented in the state of the art literature from \cite{zou2013contrastive}, outlined in Remark \ref{ossM123}. In this section we compare, using synthetically generated data, how well the two different methods recover $M_2$ and $M_3$ as a function of the sample size. To perform this experiment, we generated a set of $1000$ synthetic corpora according to the single topic model described in Section~\ref{singletopic}, with different sizes (the number of texts for each corpus); the smallest corpus contained 100 texts, the largest 10000; each text contained a random number of words, from a minimum of 3 to a maximum of 100. For each corpus, the values of the unknowns $(M, \Omega)$ have been randomly generated, and from them we have been able to obtain the theoretical values of $M_2$ and $M_3$ using equations \eqref{M2} and  \eqref{M3} and to compare those values with the one empirically estimated from data using the equations in Theorem \ref{M123Teo} for the presented method and the method from \cite{zou2013contrastive} for the competing one. Results appear in Figure~\ref{FigResultsM12}, where we show how the estimated $M_2$ and $M_3$, say  $\tilde{M_2}$ and $\tilde{M_3}$, approach the theoretical values; in particular, in the chart are represented the errors
$$
Err_2 = ||\tilde{M_2}-{M_2}||_F ,\,\,\,\,\,\,Err_3 = ||\tilde{M_3}-{M_3}||_F 
$$
as a function of the sample size $N$ used to find $\tilde{M_2}$ and $\tilde{M_2}$.
 We can see that the method of Theorem \ref{M123Teo} outperforms the state of the art technique; this is due to the fact that it gives more weight to the longer documents, where the signal is more clear, and less to the shorter, where the signal is noisier.

\subsection{Recovering \texorpdfstring{$(M,\Omega)$}{} from a random sample}

In this section we want to test the ability of the algorithm presented in this paper to recover the unknown parameters from random set of data, comparing it with state of the art methods. We will perform two experiments, one analyzing the reconstruction accuracy, and another one studying the stability of the results.
\subsubsection{Reconstruction Accuracy}
We fix a dictionary of $n = 100$ words with $k=5$ topics and we proceed as in the previous section to generate the sample $X$, distributed as a Single Topic Model: for various sizes comprised between $N=50$ and $N = 1000$ we generate synthetic corpora and we use them to learn the model parameters. For each sample corpus $X$ we proceed as follows:
\begin{itemize}
\item We estimate the values of  $\tilde{M_1},\tilde{M_2}$ and $\tilde{M_3}$ using Theorem \ref{M123Teo}.
\item We retrieve from the estimated $\tilde{M_1},\tilde{M_2}$ and $\tilde{M_3}$ the pair of unknowns $(\tilde{M},\tilde{\Omega})$ using SVTD as in Algorithm \ref{alg2}. Also we generate an alternative solutions using the decomposition algorithms from \citet{TensorLatent} (''Tensor power method''), from \citet{SpectralLatent} (''Eigendecomposition method'') and from \citet{SpectralLDA} (''SVD method''), that are the current reference methods. 
\item Each time we generate a solution, we register the time in seconds employed by the various algorithms. For each method, we represent the average time employed to perform the parameters recovery along the various runs in Figure \ref{fig:time}.
\item For each set of retrieved parameters $(\tilde{M},\tilde{\Omega})$ we calculate the learning error as follows:
$$
Err = ||\tilde{M}diag(\tilde{\Omega})\tilde{M}'-{M}diag({\Omega})M'||_F
$$
where $(M,\Omega)$ are the parameters used to generated the random sample corpus.
\item We plot in Figure \ref{FigResultsSplit} the results of the analysis as a function of $N$.
\end{itemize}
 \begin{figure}[h!]
    \centering
  \begin{subfigure}[b]{0.65\textwidth}
            \centering
            \includegraphics[height=5cm]{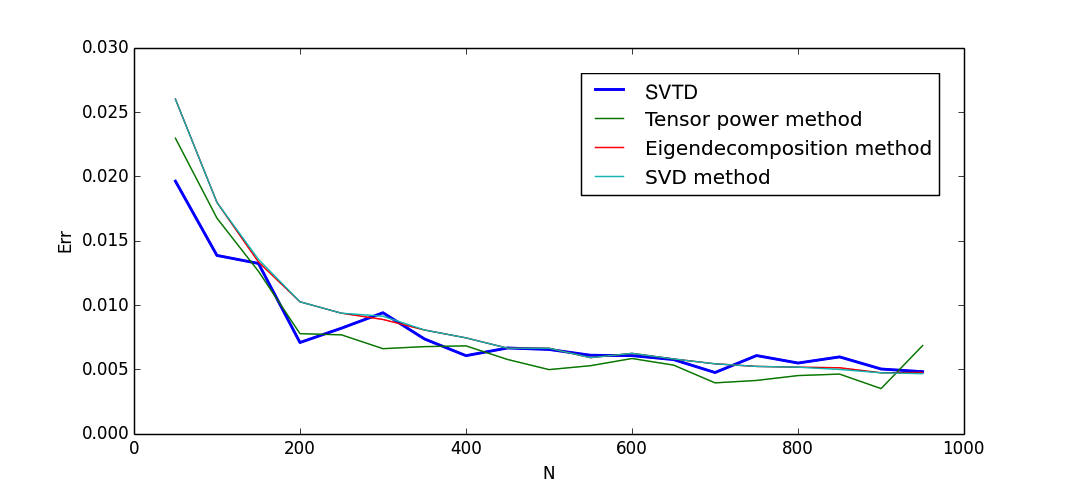}
            \caption{}
\label{FigResultsSplit} 
\end{subfigure}
    \begin{subfigure}[b]{0.33\textwidth}
            \centering
            \includegraphics[height=5cm]{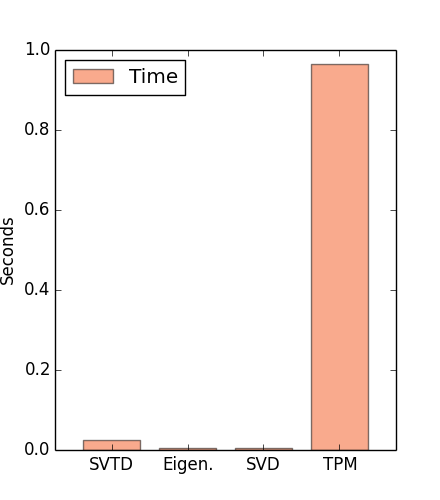}
            \caption{}
            \label{fig:time}
    \end{subfigure}

\caption{The $x-$axis of the Figure \ref{FigResultsSplit} represents the size of the synthetic text corpora, wile the $y-$axis is $Err$, the reconstruction error for the various tested methods: SVTD (blue line) behaves in a very similar way to tensor power method. In Figure \ref{fig:time} the running times are represented; we can see that TPM has sensibly longer running time, due to the worst dependence on the number of latent states, while matrix methods behave similarly.}
\end{figure}
First of all we can see that all the methods perform similarly; peaks are present when a method provides results that are far from the latent variable used to generate the sample. The performances of SVTD seems to be comparable with those of TPM, and in particular it seems to be less sensitive to noise than the two simultaneous diagonalization methods, as it performs better when the number of texts is small. In Figure \ref{fig:time}, the average running times of the various methods are presented. As expected, matrix-based methods preform similarly, and work faster than TPM, as a consequence of the better dependence on the number of latent states. SVTD has a slightly larger running time, due to the feature selection process outlined in Remark \ref{selectr}.
\subsubsection{Stability}
We now analyze a second scenario; we generate a random sample $X$ as before, with $N = 50$ texts; then we incrementally add to this corpus new texts, until when we reach $N = 200$ total texts. So, at each step $N$, we add a text to the corpus, we recalculate the tensors to be decomposed, $\tilde{M_1},\tilde{M_2}$ and $\tilde{M_3}$, and from them the model parameters $(\tilde{M}_N,\tilde{\Omega}_N)$, with the four methods described before. Then we calculate the variation of the parameters as
$$
Var_N = ||\tilde{M}_N-\tilde{M}_{N-1}||.
$$
The scope of this experiments is to understand how small variations in the inputs affect the final results. The more stable is a method, the smaller will be the registered variations between a iteration and the next. As the Eigendecomposition and SVD methods both rely on randomized vectors, here we fix those vectors, and we use them for all the tests, in order to increase their stability. Results can be seen in Figure \ref{Fig:stab}.
\begin{figure}[tb]
  \centering
\includegraphics[height=5cm]{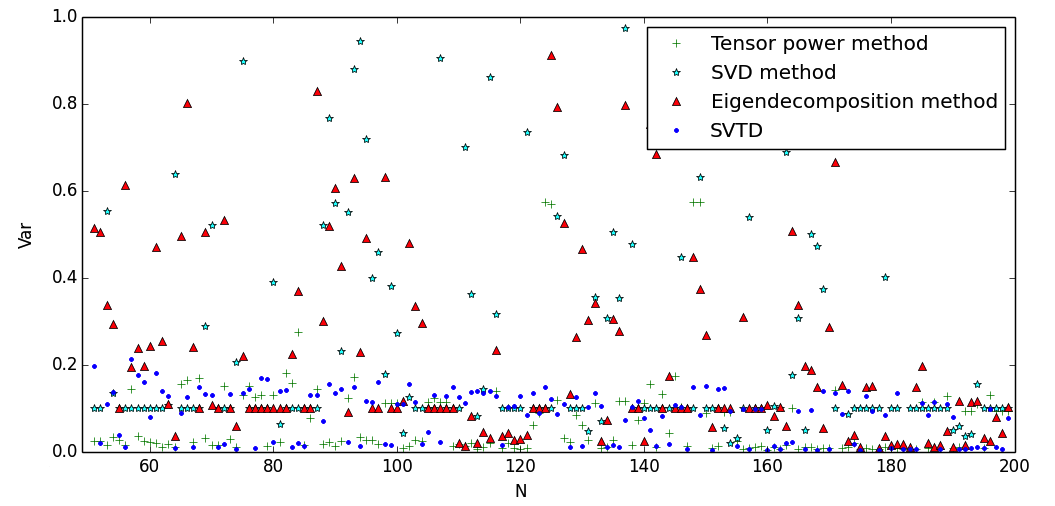}
\caption{The $x-$axis of the figure represents the size of the synthetic text corpora, wile the $y-$axis is $Var$ for the various tested methods: SVTD (blue points) proves to be much more stable than the other simultaneous diagonalization methods.}\label{Fig:stab} 
\end{figure}
As expected, SVTD provides a superior stability with respect to the other two simultaneous diagonalization methods. Indeed, we can see that the variations on the provided parameters are always very small, if compared with the red and cyan dots (the competing matrix-based methods). This might be attributed to the fact that we are not using directly the singular vectors, as the competing methods, but we rely on the singular values. Instead, the stability of SVTD in this experiments seems to be comparable to that of TPM.

\subsubsection*{Practical considerations.} 

In linear algebra operations, the running time is highly influenced by many implementation details, such as the usage of vectorized operations. All the algorithms have been implemented in Python 2.7, using \textit{numpy} \citep{van2011numpy} library for linear algebra operations. All the experiments have run on a MacBook Pro, with an Intel Core i5 processor.  

\subsection{Real data}

To perform experiments with real data, we analyzed two different corpora: the list of State of the Union addresses since 1945 to 2005 and Dante's ``Divina Commedia''. In both cases, we used a dictionary of $n=3000$ words, so the data matrix $X$ was a $N \times 3000$, where $N$ was the size of the corpus (we had $N = 65$ on the first example and $N = 100$ in the second), while tensor $M_3$ belonged to $\R^{3000\times 3000\times 3000}$. We deliberately used corpora with $N<<n$, in order to test the ability of our algorithm to work with small sets of documents. 

\subsubsection{Dante's Divina Commedia}\label{sec:DanteLda}
Dante's ``Divina Commedia'' \citep{alighieri1967commedia} is an Italian epic poem written in the first half of 14th century\footnote{The full text can be found here: \url{http://www.gutenberg.org/files/1012/1012-0.txt}}; it deals with the imaginary trip of the main character, Dante, in the afterlife, guided by Virgilio, the famous Latin poet, and Beatrice, a Florentine woman that inspired most of Dante's works. The story line represents an allegorical description of death soul's journey towards God according to medieval world view. It begins with Dante's travel trough the ``Inferno'' (Hell), where damned souls are deemed to eternal punishment according to their sins; the journey then moves to ``Purgatorio'', a seven level mountain, where, at each level, a capital sin (sins less serious than those punished in Hell) is allegorically described; here souls are discounting their punishment, before finally move to ``Paradiso'', Heaven, that is visited by Dante in the last third of the book. The book is made of 100 different chapters: 34 for Hell, 33 for Purgatory and 33 for Heaven. 
\\ \\
\textbf{Single Topic Model}\\
We run SVTD for the Single Topic Model on the ``Divina Commedia'' corpus, on the $N=100$ texts, represented by the 100 chapters of the book. We tested various possible number of topics, but surprisingly almost always the algorithm produced two significant topics. In the table \ref{Dante} we can see the results of the algorithm run with $k=2$: for each of the two topics a cluster has been defined; the most representative words are represented together with the chapters assigned to each cluster. We can see that most of the chapters of Hell are assigned to the same cluster; also the chapters of Heaven are all assigned to the same cluster while Purgatory is assigned in part to cluster 1 and in part to cluster 2.

\begin{table}[h]
\begin{tabular}{C{3cm} L  C{6cm}}\toprule
Cluster & Most representative words & Assigned chapters  \\\midrule
1 & fosso, bolgia, scoglio, avante, coda, grido, rotta, ponte, stanchi, maestro & Inferno (1,3-10, 12-34)
, Purgatorio (1-8,  10-12,20-24,26,27,29)     \\\midrule
2 & milizia, segue, intende, letizia, conosce, lumi, ama, Beatrice, piacer, cristo& Inferno (2,11)  
Paradiso (1-33), 
Purgatorio (6, 13-19, 25, 28, 30-33)  \\\midrule
 \\ 
\end{tabular}
\caption{The cluster assignment for ``Divina Commedia'' chapters.}\label{Dante}
\end{table}

\noindent
\smallskip
\\ \\
\textbf{Latent Dirichlet Allocation}\\
As a second step, we used the same corpus, with the same vocabulary, to infer for each chapter how much it deals with topic 1 and how much it deals with topic 2, using Latent Dirichlet Allocation. In particular, we retrieved from the data the tensors~$\tilde{M_2^{\alpha}}$ and~$\tilde{M_3^{\alpha}}$ from Theorem~\ref{teoLDA} and we used them to feed Algorithm~\ref{alg2}. We had to specify a value for~$\alpha_0$, that was set to~2. With Algorithm~\ref{alg2} we retrieved a pair~$(M,\alpha)$. 
We then used this pair and partially collapsed Gibbs Sampling to infer the topic mixture for each chapter. As we just have two topics, it is easy to plot the results of that inference, 
see Figure~\ref{FigResultsDante}. In this chart, the $x-$axis represents the progression of the chapters along the book, while red and blue lines represent the value of the topic proportions for each chapter. We can see that most of the first 34 chapters have a strong predominance of the first topic, marked with the red line, that consequently can be identified with the Hell topic. In the same way, second topic, or Heaven, is very strong in the last 33 chapters. 
Purgatory, in the middle part of the plot, has a mixed belonging. The amazing fact is that the proportion of the Heaven topic seems to increase as the chapters approach 
the Heaven section, corresponding to Dante's ascent of the Purgatory mountain. 

\begin{figure}[h]
  \centering 
\includegraphics[width=0.8\textwidth]{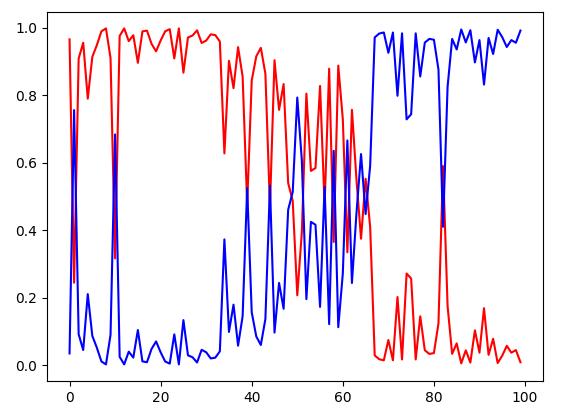}
\caption{In the figure, the $x-$axis represents the chapter of the book: the first 34 are Hell, 
from 35 to 67 we have Purgatory and the last 33 are Heaven. The red line represents how much each chapter belongs to the first cluster, while the blue line how much it belongs to second cluster.} \label{FigResultsDante}
\end{figure}
  
\subsubsection{State of the Union addresses}

Each year, the president of United States of America presents a speech to a joint session of the United States Congress, where he outlines his governative agenda, the national priorities and legislative projects. We considered the set of $N = 65$ state of the union addresses presented between 1945 and 2005, and we applied to this corpus the algorithm for the Single Topic Model, with the purpose of finding the most representative topics of the corpus and clustering the various speeches according to the learned topics\footnote{The full corpus can be found
in link \url{http://www.nltk.org/nltk_data}.}. We run SVTD assuming to have $k=5$ different topics, although with different values of $k$ results were similar; we then grouped the speeches assigning them to the topic with the highest likelihood, using the Bayesian posterior assignment of Remark \ref{ossNBMPrior}. In Table \ref{StateUnion} we can see the results of this operation. For each topic a cluster has been defined; the most representative words are represented together with the speeches assigned to each cluster; for each president, the brackets indicate the year of the speech.

\begin{table}[th!]
 \begin{tabularx}{\textwidth}{C{3cm} L C{6cm}}\toprule
Cluster & Most representative words & Assigned speeches  \\\midrule
  1 & construction, fiscal, legislative, peacetime, facilities, recommendations, projects, existing, transportation, veterans 
&  Truman (1946 to 1950), Eisenhower (1953 to 1957, 1959), Kennedy (1961 to 1963), Johnson (1966,1967), Nixon (1973), Ford (1975,1977) \\\hline
  2 & reducing, regulations, recovery, taxpayers, market, bills, weeks, nothing, gone, productivity
 & Johnson (1963 to 1965, 1968, 1969), Nixon (1970 to  1972, 1974), Ford (1976), Carter (1978), Reagan (1981 to 1988), Bush (1990, 1992) \\\hline 
 3 & ideals, soviet, missile, potential, missiles, world, conflict, struggle, countries, threat
&  Truman (1945, 1951), Eisenhower (1958, 1960), Carter (1979, 1980), Bush (1991a,1991b), G.W. Bush (2003)\\\hline
  4 & companies, invest, 21st, teachers, parents, revolution, lowest, challenge, credit, bipartisan
 & Bush (1989), Clinton (1993 to 2000) \\\hline
  5 & September, enemies, terror, compassion, terrorists, Afghanistan, relief, retirement, Iraq,  dangerous
& G.W. Bush (2001a, 2001b, 2002, 2004, 2005) \\\hline
\end{tabularx}
\caption{The cluster assignment for the State of the Union addresses.}\label{StateUnion}
\end{table}
Cluster 1 contains only speeches of the Cold War period. Words like 'peacetime', 'construction' and 'projects' characterize, for instance, the period after the second World War, with the Marshall plan and Truman doctrine. Cluster 2 has speeches that belong to a wider set of dates, but all seem to be about internal politics and economics; among the most representative words we can find 'taxpayers', 'regulations' and 'productivity', key themes of the Reagan administration. Cluster 3 is clearly related to war with words as 'missile', 'conflict' and 'struggle'; the speeches come from presidents involved in important wars (WWII and Gulf wars). Cluster 4 mainly contains Bill Clinton speeches and has words that characterize the economic expansion of the 90ies ('companies', 'invest', 'credit'). Cluster 5 has among the top words 'terror', 'Afghanistan' and 'Iraq'; they all reveal  post 9/11 politics carried on by G.W. Bush.

\section{Conclusion and Future Work}\label{sec:conclusion}

We described a simple algorithm to learn latent variable models in polynomial time which shares many good characteristics of previous spectral methods, having at least one advantage over each of them (be in efficiency, or being deterministic and more stable); together with this, we have introduced an efficient method for estimating the symmetric tensors of the moments for the Single Topic Model and for LDA. A natural future work is to adapt this algorithm to an on-line, streaming environment \citep{PCA1,PCA2}. In the theoretical front, we want to improve the perturbation Theorem \ref{pert}, removing the dependence from $\alpha$ and $\alpha_M$. In the applications side, we are interested in applying this algorithm to learn LVM in the healthcare analytics field, for instance to construct disease progression models and patient clusterings. Genetic data, where e.g. one typically has many more
genes or SNPs than sequenced individuals, would also be of interest. 

\subsection*{Acknowledgements}We are grateful to thank Daniel Hsu for his guidance on the spectral method for LDA. M.~Casanellas is is partially funded by AGAUR project 2014 SGR-634 and MINECO/FEDER project MTM2015-69135-P. R.~Gavald\`a
is partially funded by AGAUR project 2014 SGR-890 (MACDA) 
and by MINECO project TIN2014-57226-P (APCOM).

\newpage

\appendix
\section*{Appendix}
\section{Proofs for Section 2}
\subsection{Proof of Theorem 2.1}
\proof
We will prove the statements only for 
$$
\E(\frac{\sum_{i=1}^{N}(X^{(i)})_h(X^{(i)})_l}{\sum_{i=1}^{N}(c_i-1)c_i}) =   \sum_{j=1}^{k}\omega_j\mu_{h,j}\mu_{l,j}.
$$
Similar arguments hold for the other equations. It is easy to see, by conditional independence, that 
$$
E((X^{(i)})_h(X^{(i)})_l) = \sum_{j=1}^{k}\omega_jE((X^{(i)})_h(X^{(i)})_l|Y=j)   
$$
but the conditioned $(X^{(i)})_h$ and $(X^{(i)})_l$ are components of a multinomial distribution and so
$$
 \sum_{j=1}^{k}\omega_jE((X^{(i)})_h(X^{(i)})_l|Y=j) = \sum_{j=1}^{k}\omega_j(c_i^2-c_i)\mu_{h,j}\mu_{l,j}
$$
which implies the thesis. 
\endproof
\subsection{Proof of Theorem 2.2}
\proof
We want to express in a suitable way the elements of the matrix
\begin{equation}\label{eq:diffm2}
\tilde{M_2}-M_2
\end{equation}
and then express a bound using McDiarmid's inequality \citep{mcdiarmid1989method}.
We know by construction that, for any $i\in\{1,...,N\}$ it holds that
$$
X^{(i)} = \sum_{j=1}^{c_i}x_j^{(i)}
$$
where each $x_j^{(i)}$ is the $j-$th word of the document. 
We thus consider the set of all the words from all the documents:  
$$
\mathcal{X} = (x^{(1)}_1,...,x^{(1)}_{c_1},...,x^{(N)}_1,...,x^{(N)}_{c_N}).
$$
It is easy to see that $\tilde{M_2}$ can be expressed as a function of $ \mathcal{X}$, for all pairs $u,v \in \{1,...,n\}$ we have
$$
(\tilde{M_2})_{u,v} (\mathcal{X}) = \frac{\sum_{i\neq j}(x^{(1)}_i)_u(x^{(1)}_j)_v + ... +
 \sum_{i\neq j}(x^{(N)}_i)_u(x^{(N)}_j)_v}{C_{2}}
$$
where
$$
C_{2}=\sum_{i=1}^{N}c_i(c_i-1).
$$
We now define the following function:
$$
\Phi(\mathcal{X}) = || \tilde{M_2}(\mathcal{X})-M_2 ||_F
$$
and observe that, given
$$
\mathcal{X} = (x^{(1)}_1,...,x^{(1)}_{c_1},...,x^{(l)}_{i},...,x^{(N)}_1,...,x^{(M)}_{c_N})
$$
and
$$
{\mathcal{X}}' = (x^{(1)}_1,...,x^{(1)}_{c_1},...,{x^{(l)}_{i}}',...,x^{(N)}_1,...,x^{(M)}_{c_N})
$$
we have
$$
|\Phi(\mathcal{X}) - \Phi({\mathcal{X}}')|\leq || \tilde{M_2}(\mathcal{X})-\tilde{M_2}({\mathcal{X}}') ||_F =
$$
$$
= \sqrt{\sum_{u,v = 1}^{n} \left( {\frac{\sum_{i\neq j}(x^{(l)}_j)_u((x^{(l)}_i)_v-({x^{(l)}_i}')_v)}{C_{2}}}\right)^2} \leq \frac{\sqrt{2}\max_j(c_j)}{C_{2}}.
$$
We are now able to apply McDiarmid's inequality stating that
$$
\P(||\tilde{M_2}-M_2||_F>\E(||\tilde{M_2}-M_2||_F) + \epsilon) \leq  e^{-\frac{\epsilon^2C_{2}^2}{(\max_j(c_j))^2C_1}}. 
$$
So, by setting
$$ 
t = \frac{\epsilon C_2}{\max_j(c_j)\sqrt{C_1}}
$$
we get
$$
\P(||\tilde{M_2}-M_2||_F> \E(||\tilde{M_2}-M_2||_F) +  t\frac{\max_j(c_j)\sqrt{C_1}}{ C_2} ) \leq   e^{-t^2}.
$$
We now provide a bound for $\E(||\tilde{M_2}-M_2||_F)$. 
We begin observing that 
$$
\tilde{M_2} = \sum_{i=1}^{N}w_i \tilde{M_2}^{(i)}
$$
where $
w_i = \frac{c_i(c_i-1)}{C_2}
$ and $\tilde{M_2}^{(i)}$ are independent matrices defined as follows:
$$
(\tilde{M_2}^{(i)})_{(u,v)}  =  \frac{\sum_{l\neq j}(x^{(i)}_l)_u(x^{(i)}_j)_v }{c_i(c_i-1)}
$$
Notice that, for any $i$, 
$
\E(\tilde{M_2}^{(i)}) = M_2.
$
{Using Jensen's inequality we have}
$$
\E(||\tilde{M_2}-M_2||_F) \leq \sqrt{\E(||\tilde{M_2}-M_2||_F^2)}.
$$
This last term is equal to
$$
  \sqrt{\E(||\tilde{M_2}||_F^2)-||M_2||_F^2} = 
\sqrt{\sum_{u,v}\E((\sum_{i=1}^{N}w_i(\tilde{M_2}^{(i)})_{(u,v)})^2) -||M_2||_F^2} = 
$$
$$
=   \sqrt{\sum_{u,v}\E(\sum_{i=1}^{N}w_i^2(\tilde{M_2}^{(i)})_{(u,v)}^2) + \sum_{u,v}\E(\sum_{i\neq j}w_jw_i(\tilde{M_2}^{(i)})_{(u,v)}(\tilde{M_2}^{(j)})_{(u,v)}) -||M_2||_F^2} 
$$
and using the fact that $\E(\tilde{M_2}^{(i)}_{(u,v)}\tilde{M_2}^{(j)}_{(u,v)}) = (M_2)_{(u,v)}^2$, this equals
$$
\sqrt{\sum_{u,v}\sum_{i=1}^{N}w_i^2\E(||\tilde{M_2}^{(i)}||_F^2) + \sum_{i\neq j}w_jw_i||M_2||_F^2  -||M_2||_F^2} \, .
$$
Now using that $||\tilde{M_2}^{(i)}||_F\leq 1$, we can bound this from above by
$$
\sqrt{\sum_{i=1}^{N}w_i^2 + ||M_2||_F^2 (\sum_{i\neq j}w_jw_i -1)} = 
\sqrt{\sum_{i=1}^{N}w_i^2(1 - ||M_2||_F^2)}.
$$
where in the last equality we used the fact that $\sum_{i\neq j}w_jw_i = 1-\sum_{i=1}^{N}w_i^2$.
So, if we call $ W_2^{(N)} = \sum_{i=1}^{N}w_i^2$, we have
 $
\E(||\tilde{M_2}-M_2||_F) \leq \sqrt{W_2^{(N)} (1 - ||M_2||_F^2)},
$
from which we obtain
$$
\P(||\tilde{M_2}-M_2||_F> \sqrt{W_2^{(N)} (1 - ||M_2||_F^2)} +  t\frac{\max_j(c_j)\sqrt{C_1}}{ C_2} ) \leq   e^{-t^2}.
$$
In conclusion, we can state that if 
 $
e^{-t^2} = \delta 
$
we get, for any $\delta \in (0,1]$
$$
\P(||\tilde{M_2}-M_2||_F> \epsilon ) \leq   \delta 
$$
where
$$
 \epsilon = \sqrt{W_2^{(N)} (1 - ||M_2||_F^2)} + \sqrt{\log(\frac{1}{\delta}} )\frac{\max_j(c_j)\sqrt{C_1}}{ C_2}.
$$
A similar argument works for $M_3$.
\endproof

\section{Proofs for Section 4}
 
 \subsection{Proof of Lemma 4.1}
\proof
We will use the same notation of the proof of Theorem \ref{teoalgo}. 
We consider any $E$ with rank $k$ satisfying 
$$
M_2 = EE'
$$
and we recall that we can find an isometry $O$ such that 
\begin{equation}\label{TrueEO}
EO = M (diag(\Omega))^{\frac{1}{2}}
\end{equation}
Now, given a feature $r$, we can easily define $E_r \in \R^{(n-1)\times k}$ as $E$ with the $r-th$ row removed.
It is easy to see that 
$$
M_{2,r} = E_rE_r'
$$
and that there exist an isometry  $O_r$ realizing the equation  
$$ 
E_rO_r = M_r (diag(\Omega))^{\frac{1}{2}}.
$$
We will prove that, for each $r$, we have $O = O_r$.
First, we define the following notation:
$$
\M_r := M_r (diag(\Omega))^{\frac{1}{2}}
$$
$$
\M := M (diag(\Omega))^{\frac{1}{2}}.
$$
Also, it will be useful to define the rows of $E$ and $\M$:
$$
E = \begin{pmatrix}
  e_1 \\
  \vdots \\
  e_n
 \end{pmatrix} \in \R^{n\times k},\,\,\,\,\,
\M = \begin{pmatrix}
  m_1\\
  \vdots \\
  m_n
 \end{pmatrix} \in \R^{n\times k}.
$$
We will prove the theorem only for $r = n$, but the proof is very similar for the other cases.
By equation \eqref{TrueEO} we have that
$$
O = E^{\star}\M = (E'E)^{-1}E'\M =
$$
$$
 = (E_r'E_r+e_r'e_r)^{-1}(E_r'\M_r + e_r'm_r).
$$
Now, using the Sherman–Morrison formula \citep{sherman1950adjustment} we obtain
$$
O = \big[(E_r'E_r)^{-1}+\frac{(E_r'E_r)^{-1}e_r'e_r(E_r'E_r)^{-1}}{1+e_r(E_r'E_r)^{-1}e_r'}\big](E_r'\M_r + e_r'm_r)
$$
and so
$$
O = (E_r'E_r)^{-1}E_r'\M_r + (E_r'E_r)^{-1}e_r'm_r -
$$
$$
-\frac{(E_r'E_r)^{-1}e_r'e_r(E_r'E_r)^{-1}E_r'\M_r + (E_r'E_r)^{-1}e_r'e_r(E_r'E_r)^{-1}e_r'm_r}{1+e_r(E_r'E_r)^{-1}e_r'} = 
$$
$$
 = O_r + \frac{(E_r'E_r)^{-1}e_r'm_r - (E_r'E_r)^{-1}e_r'e_r(E_r'E_r)^{-1}E_r'\M_r}{1+e_r(E_r'E_r)^{-1}e_r'}.
$$
We now just have to prove that
$$
m_r = e_r(E_r'E_r)^{-1}E_r'\M_r = e_rE_r^{\star}\M_r.
$$
We know that
$$
M_2 = EE' = \M\M'
$$
thus
$$  
\begin{pmatrix}
  E_r \\
  e_r
 \end{pmatrix}
\begin{pmatrix}
  E_r' &  e_r'
 \end{pmatrix}
 =
\begin{pmatrix}
  \M_r \\
  m_r
 \end{pmatrix} 
\begin{pmatrix}
  \M_r' &  m_r'
 \end{pmatrix}
$$
so the following system holds:
$$
\begin{cases} E_rE_r' = \M_r\M_r' \\ e_rE_r' = m_r\M_r'
\end{cases}
$$
so
$$
E_r' = E_r^{\star}\M_r\M_r'
$$
from which we have our thesis
$$
m_r\M_r' = e_rE_r^{\star}\M_r\M_r' \Longrightarrow m_r  = e_rE_r^{\star}\M_r.
$$
\endproof

\section{Proofs for Section 5}

\subsection{Proof of Theorem 5.1}

\proof
The goal of the proof is to develop a perturbation bound for each row $m_i$ of the unknown matrix {$M$ }
such that, 
$
||\tilde{m}_i-{m}_i||_2 \leq  Bound(\epsilon),
$ {for a certain function  $Bound(\epsilon)$.}
We notice, from Algorithm \ref{alg2}, step 8, that each $\tilde{m}_i$ is obtained as the diagonal entries of the following matrix:
$$
\tilde{O'}\tilde{E}_i^{\star} \tilde{M}_{3,i} (\tilde{E_i'})^{\star}\tilde{O}
$$
and so, we will need to find the perturbations of the matrices composing this equation, as the following relation holds:
$$
||\tilde{m}_i-{m}_i||_2\leq|| \tilde{O'}\tilde{E}_i^{\star} \tilde{M_{3,i}} (\tilde{E_i'})^{\star}\tilde{O} - {O'}{E}_i^{\star} {M_{3,i}} ({E_i'})^{\star}{O}||_F.
$$
In short, having perturbation bounds on $\tilde{M}_{3,i}$ ,$\tilde{E}_i^{\star}$ and $\tilde{O}$ will be sufficient to reach our goal.

\subsubsection*{Perturbations on $\tilde{M}_{3,i}$}

We know, by hypothesis of the theorem, that
$$
||\tilde{M}_{3,i}-{M_{3,i}}||_F = ||\Delta_{M_{3,i}}||_F  < \epsilon.
$$ 
 
\subsubsection*{Perturbations on $\tilde{E}_i^{\star}$}

It is a known fact \citep[see][]{SingVal} that, given the SVD  
$$
\tilde{M_2} = \tilde{U}\tilde{S}\tilde{V}',\,\,\,\,\,M_2 = {U}{S}{U}',
$$
if
$$
||\tilde{M_2} -{M_2} ||_F <\epsilon,
$$
we have that
\begin{equation}\label{SErr}
||\tilde{S}-S||_F \leq  \epsilon.
\end{equation}
Algorithm \ref{alg2} considers at step 2 the following approximation of $E$
$$
\tilde{E} = \tilde{U}_k(\tilde{S}_k)^{\frac{1}{2}},
$$
while the unperturbed value of $E$ can be found as 
$$
{E} = {U}_k({S}_k)^{\frac{1}{2}},
$$
where the subscript $k$ indicates the truncation at the $k-th$ singular value.
So, to reach a perturbation bound on $\tilde{E}_i^{\star}$ for a given $i$, we first need to look for a perturbation bound on $\tilde{E}$, that  will be obtained bounding the error of $(\tilde{S}_k)^{\frac{1}{2}}$ and $\tilde{U}_k$. The first one is a consequence of equation \eqref{SErr}: {if $\Delta_S =(\tilde{S}_k)^{\frac{1}{2}}-S_k^{\frac{1}{2}}$, we have}
$$
||\Delta_S ||_F <\frac{\epsilon}{2 \sqrt{\sigma_{k}(M_2)}},
$$
where $\sigma_{k}(M_2)$ is the $k-$th the singular value of $M_2$.
To find a bound on $\tilde{U}_k$ we will use Lemma \ref{lemma:singvect} to get
$$
||\tilde{U}_k-U_k||_F< \sqrt{8k}\frac{2\epsilon ||M_2||^{2}_{F}  + \epsilon^2}{\alpha_{M_2}} .
$$
where 
$$
\alpha_{M_2} =\min_{i\leq k}{(\sigma_{i}(M_2)^2-\sigma_{i+1}(M_2)^2)}
$$
and $\sigma_{i}(M_2)$ are the singular values of $M_2$.

We thus conclude that, if $\Delta_U=\tilde{U}_k-U_k$,
$$
\tilde{E} = E + \Delta_U S^{\frac{1}{2}} + U \Delta_S  +\Delta_U\Delta_S
$$
and hence
$$
||\tilde{E}- E||_F<f(\epsilon) = \epsilon \left( ||S^{\frac{1}{2}}||_F\sqrt{8k}\frac{2 ||M_2||^{2}_{F}  + \epsilon}{\alpha_{M_2}}+  \frac{ ||U||_F}{2 \sqrt{\sigma_{k}(M_2)} } +  \sqrt{8k}\frac{   2\epsilon||M_2||^{2}_{F}  + \epsilon^2 }{\alpha_{M_2}(2 \sqrt{\sigma_{k}(M_2)})}\right).
$$
We are now ready to find a perturbation bound on the pseudoinverse of $\tilde{E}$ after the removal of row $i$.
This can be accomplished using a known bound from \citep{PseudoInv}:
if
$$
||\tilde{E_i}- E_i||_F<f(\epsilon) 
$$
then
$$
||\tilde{E_i}^{\star}- E_i^{\star}||_F\leq f(\epsilon)\tau(E)
$$
where 
$$
\tau(E) = ||E^{\star}||_F^2 + ||(E'E)^{-1}||_F||\mathbb{I} - EE^{\star}||_F.
$$
\subsubsection*{Perturbations on $\tilde{O}$}

We now look for the value of $g(\epsilon) = ||O- \tilde{O}||_F$. In particular, $O$ comes from the decomposition of $E_r^{\star} M_{3,r} (E_r')^{\star}$:
\begin{equation}\label{eq:O}
E_r^{\star} M_{3,r} (E_r')^{\star} = O diag((\mu_{r,1},...,\mu_{r,k}))O'.
\end{equation}  
while $\tilde{O}$ is the set of the left singular vectors of
$
\tilde{E_r^{\star}} \tilde{M_{3,r}} (\tilde{E_r}')^{\star}.
$
First, we observe that 
\begin{multline}
||E_r^{\star} M_{3,r} (E_r')^{\star}  -\tilde{E_r^{\star}} \tilde{M_{3,r}} (\tilde{E_r}')^{\star}||_F \leq 
\\
\leq 2f(\epsilon)\tau(E) ||E^{\star}||_F||M_{3,r}||_F +  \epsilon||E^{\star}||_F^2  +\\ +2\epsilon f(\epsilon)\tau(E) ||E^{\star}||_F +
f(\epsilon)^2\tau(E)^2(||M_{3,r}||_F + \epsilon) = h(\epsilon).
\end{multline}
Using  Corollary  \ref{cor:chenSVD}, we assume the hypothesis that
\begin{equation}\label{eq:pertBound2}
h(\epsilon) \leq\frac{\alpha_{r}^2}{\sqrt{2}\left(2||{H_r}||_F(1+\sqrt{1 - \frac{1}{k}}) + \sqrt{\alpha_{r}^2 + 4||H_r||_F^2(1+\sqrt{1 - \frac{1}{k}})^2}\right)}
\end{equation} 
where ${H_r} = E_r^{\star} M_{3,r} (E_r')^{\star} 
$ and
$
\alpha_{r} = \min_{i\neq j}(|\mu_{r,i}-\mu_{r,j} |)
$, to get that 
$$
||\tilde{O}- O||_F< 2\sqrt{2}\frac{h(\epsilon)}{\alpha_{r}} = g(\epsilon).
$$
We are now able to conclude our proof by analyzing
$$
||m_i-\tilde{m_i}||_2\leq|| \tilde{O'}\tilde{E}_i^{\star} \tilde{M_{3,i}} (\tilde{E_i'})^{\star}\tilde{O} - {O'}{E}_i^{\star} {M_{3,i}} ({E_i'})^{\star}{O}||_F \leq
$$
$$
\leq  P_1 g(\epsilon) + P_2 \epsilon + P_3 f(\epsilon)  + O(\epsilon^2P_4),
$$
where $P_1,P_2$ and $P_3$ are polynomials in  $ ||E_i||_F,||O||_F$ and $||M_{3,i}||_F$, and $P_4$ is a polynomial in  $ ||E_i||_F,||O||_F,||M_{3,i}||_F,g(\epsilon), \epsilon$ and $ f(\epsilon)$ . The thesis follows making explicit these polynomials.

\endproof
\begin{oss}\label{oss:numberbound}\rm
In the statement of Theorem \ref{pert}, we said that there exists a number $\gamma(M,\Omega)$, such that, if $\epsilon<\gamma(M,\Omega)$, the perturbation bound of the thesis works. Looking at the proof of the theorem, we are able to explicitly calculate this number, just by solving the inequality  \eqref{eq:pertBound2}. We can practically think at $\gamma(M,\Omega)$ as the largest value of $\epsilon$ that satisfies this inequality.
\end{oss}

\begin{lemma}\label{lemma:singvect}
Consider $M_2$, the perturbed $\tilde{M}_2$, and their SVD, $
\tilde{M_2} = \tilde{U}\tilde{S}\tilde{V}'$, $M_2 = {U}{S}{U}'$.
Let $\tilde{U}_k$ and ${U}_k$ be matrices of the first $k$ left singular vectors of $\tilde{M_2}$ and  ${M_2}$.
Define
$$
\alpha_{M_2}=\min_{i\leq k}{(\sigma_{i}(M_2)^2-\sigma_{i+1}(M_2)^2)}
$$
If $||\tilde{M_2}-{M_2}||<\epsilon$, the following relation holds
$$
||\tilde{U}_k-U_k||_F< \sqrt{8k}\frac{2\epsilon ||M_2||^{2}_{F}  + \epsilon^2}{\alpha_{M_2}} .
$$
\end{lemma}
\proof
Consider 
$$
\tilde{M_2}\tilde{M_2}' = \tilde{U}\tilde{S}^2\tilde{U'},\,\,\,\,\,M_2M_2' = {U}{S^2}{U}'.
$$
Then $
||\tilde{M_2}\tilde{M_2}' - M_2{M_2}' ||_F \leq 2\epsilon ||M_2||^{2}_{F}  + \epsilon^2.
$
Take now the matrix of the first $k$ columns of $U$ and $\tilde{U}$, that are eigenvectors of ${M_2}{M_2}'$ and $\tilde{M_2}\tilde{M_2}'$, obtaining $U_k = [u_1,..,u_k]$ and $\tilde{U}_k = [\tilde{u}_1,..,\tilde{u}_k]$. From Theorem \ref{teo:useful} we have that, for any $i=1,...,k$, holds
$$
||u_i - \tilde{u}_i||\leq 2^{\frac{3}{2}}\frac{2\epsilon ||M_2||^{2}_{F}  + \epsilon^2}{\min(\sigma_{i-1}(M_2)^2 -\sigma_{i}(M_2)^2,\sigma_{i}(M_2)^2 -\sigma_{i+1}(M_2)^2  )}\leq 2^{\frac{3}{2}}\frac{2\epsilon ||M_2||^{2}_{F}  + \epsilon^2}{\alpha_{M_2}}
$$
from which the thesis follows.
\endproof
 
The following results, taken from \citet{yu2015useful} and \citet{SingVect}, present perturbation bounds on the eigenvectors and on the singular vectors of symmetric matrices. 

\begin{teo}[Cor. 1, pg. 4  \citealt{yu2015useful}]\label{teo:useful}
Consider $A$ and $\tilde{A}$ two symmetric matrices in $\R^{n\times n}$, with eigenvalues $\lambda_1 \geq ...\geq \lambda_n$ and $\tilde{\lambda}_1 \geq ...\geq \tilde{\lambda}_n$. Fix a $j\in \{1,...,n\}$ and assume that $\min(\lambda_{j-1} -\lambda_{j},\lambda_{j} -\lambda_{j+1}  )>0 $, where we define $\lambda_0 = \infty$ and $\lambda_{n+1} = -\infty$. If $v\in R^n$ (resp. $\tilde{v}$) is an eigenvector of $A$  (resp. $\tilde{A}$), associated to $\lambda_i$  (resp. $\tilde{\lambda_i}$), then
$$
||v - \tilde{v}||\leq \frac{2^{\frac{3}{2}}||A-\tilde{A}||}{\min(\lambda_{j-1} -\lambda_{j},\lambda_{j} -\lambda_{j+1}  )}
$$
\end{teo}

 \begin{teo}[Thm. 3.2,  \citealt{SingVect}]\label{teo:chenSVD}
Let $B\in \R^{k\times k}$ be a matrix, with SVD 
$$
B = U diag((\sigma_1,...,\sigma_k))V^{'}
$$
with $\sigma_1>\sigma_2>...>\sigma_k>0$, and let 
$$
\tilde{B} = B + \Delta_B
$$
be a perturbed matrix, with SVD
$$
\tilde{B} = \tilde{U} diag((\tilde{\sigma}_1,...,\tilde{\sigma}_k))\tilde{V}^{'}
$$
Define:
$$
\alpha_B = \min_{i\neq j}|\sigma_i-\sigma_j|>0,\,\,\,\,\,\,\epsilon_B = \frac{\sqrt{2}|| \Delta_B ||_F}{\alpha_B },\,\,\,\,\,\,\gamma_B = \frac{||B||_F}{\alpha_B}(1+\sqrt{1 - \frac{1}{k}})
$$
Then, if 
\begin{equation}\label{eq:epsboundchenSVD}
 \epsilon_B \leq \frac{1}{2\gamma_B + \sqrt{1 + 4\gamma_B^2}}
\end{equation}

The following upper bound holds:
\begin{equation}\label{eq:UchenSVD}
||U - \tilde{U}||_F \leq \frac{\sqrt{2}\epsilon_B}{\sqrt{1 - 2\gamma_B \epsilon_B+ \sqrt{1 - \epsilon_B^2 - 4\gamma_B\epsilon_B}}}
\end{equation}

 \end{teo}
 The following corollary is essentially a rewriting of the previous theorem.
 \begin{cor}\label{cor:chenSVD}
In the same setting of Theorem \ref{teo:chenSVD}, if there exists an $\epsilon>0$ such that
$$
||\Delta_B||<\epsilon\leq\frac{\alpha_B^2}{\sqrt{2}\left(2||B||_F(1+\sqrt{1 - \frac{1}{k}}) + \sqrt{\alpha_B^2 + 4||B||_F^2(1+\sqrt{1 - \frac{1}{k}})^2}\right)}
$$ 
then it holds
$$
||U - \tilde{U}||_F \leq 2\sqrt{2}\frac{\epsilon}{\alpha_B}
$$
\end{cor} 
  \proof {Note that if $\epsilon$ satisfies the hypothesis of the corollary, then \eqref{eq:epsboundchenSVD} is satisfied and hence we have \eqref{eq:UchenSVD}:}
 $$
 ||U - \tilde{U}||_F \leq \frac{\sqrt{2}\epsilon_B}{\sqrt{1 - 2\gamma_B \epsilon_B + \sqrt{1 - \epsilon_B^2 - 4\gamma_B\epsilon_B}}} \leq \frac{\sqrt{2}\epsilon_B}{\sqrt{1 - 2\gamma_B \epsilon_B }}
 $$
 Now we plug in the last equation the bound of \eqref{eq:epsboundchenSVD}, to get
  $$
  ||U - \tilde{U}||_F \leq  \frac{\sqrt{2}\epsilon_B}{\sqrt{1 - \frac{2\gamma_B }{2\gamma_B + \sqrt{1 + 4\gamma_B^2}} }} =   \sqrt{2}\epsilon_B\sqrt{ \frac{2\gamma_B + \sqrt{1 + 4\gamma_B^2}}{\sqrt{1 + 4\gamma_B^2}} } =   \sqrt{2}\epsilon_B\sqrt{ \frac{2\gamma_B }{\sqrt{1 + 4\gamma_B^2}} +1}\leq 2\epsilon_B
 $$
 \endproof

\vskip 0.2in
\bibliography{AR_biblio} 
\bibliographystyle{apa}

\end{document}